\documentclass[graybox]{svmult}
\pdfoutput=1

\usepackage{mathptmx}       % selects Times Roman as basic font
\usepackage{helvet}         % selects Helvetica as sans-serif font
\usepackage{courier}        % selects Courier as typewriter font
\usepackage{type1cm}        % activate if the above 3 fonts are
\usepackage{makeidx}         % allows index generation
\usepackage{graphicx}        % standard LaTeX graphics tool
\usepackage{multicol}        % used for the two-column index
\usepackage[bottom]{footmisc}% places footnotes at page bottom

\usepackage {amsmath}
\usepackage {amssymb}
\usepackage{algorithm}
\usepackage{algorithmic}
\usepackage{xspace}

\usepackage{paralist}

\newcommand{\knowref}[2][]{\ifx&#1&({\it k.\ref{#2}})\else({\it k.\ref{#1}-\ref{#2}})\fi}
\newcommand{\capref}[2][]{\ifx&#1&({\it c.\ref{#2}})\else({\it c.\ref{#1}-\ref{#2}})\fi}
\newcommand{\algoName}{density-augmented gradient descent}
\usepackage{url}

\renewcommand{\vec}[1]{\mathbf{#1}}

\newcommand{\nth}[2]{\ensuremath{{#1}^{\mbox{\scriptsize #2}}}} % 1st, 2nd, ith
\newcommand{\ie}{\emph{i.e.}}
\newcommand{\eg}{\emph{e.g.}}
\newcommand{\etal}{\emph{et al.}}
\newcommand{\etc}{\emph{etc.}}

\newcommand{\diff}[2]{\frac{\partial #1}{\partial #2}}
\newcommand{\T}{\top}
\newcommand{\R}{\mathbb{R}}

\newcommand{\tsum}{\textstyle\sum}

\newcommand{\mech}{\ensuremath{\mathcal L}\xspace}
\newcommand{\npmech}{\ensuremath{\hat{\mech}}\xspace}
\newcommand{\hypotheses}{\ensuremath{\mathcal H}\xspace}
\newcommand{\dppriv}{\ensuremath{\beta}\xspace}
\newcommand{\dpacc}{\ensuremath{\epsilon}\xspace}
\newcommand{\dpconf}{\ensuremath{\delta}\xspace}

\newcommand{\Srndic}{\v{S}rndi\'{c}}

\makeindex             % used for the subject index
\begin{document}

\title*{Security Evaluation of Support Vector Machines in Adversarial Environments}
\titlerunning{Security Evaluation of SVMs in Adversarial Environments} 
\author{Battista Biggio, Igino Corona, Blaine Nelson, Benjamin~I.~P.~Rubinstein, Davide Maiorca, Giorgio Fumera, Giorgio Giacinto, and Fabio Roli}
\authorrunning{Biggio, Corona, Nelson, Rubinstein, Maiorca, Fumera, Giacinto, Roli}
\institute{Battista Biggio, Igino Corona, Davide Maiorca, Giorgio Fumera, Giorgio Giacinto, and Fabio Roli \at Department of Electrical and Electronic Engineering, University of Cagliari,\\ Piazza d'Armi 09123, Cagliari, Italy.\\
\email{{battista.biggio,igino.corona,davide.maiorca}@diee.unica.it}\\
\email{{fumera,giacinto,roli}@diee.unica.it}\\
Blaine Nelson \at Institut f\"ur Informatik, Universit\"at Potsdam,
August-Bebel-Stra{\ss}e 89, 14482 Potsdam, Germany.\\
\email{blaine.nelson@gmail.com}\\
Benjamin I. P. Rubinstein \at IBM Research, Lvl 5 / 204 Lygon Street, Carlton, VIC 3053, Australia.\\
\email{ben@bipr.net}} % check affiliation!

\maketitle

\abstract{Support Vector Machines (SVMs) are among the most popular classification techniques adopted in security applications like malware detection, intrusion detection, and spam filtering. However, if SVMs are to be incorporated in real-world security systems, they must be able to cope with attack patterns that can either mislead the learning algorithm (poisoning), evade detection (evasion), or gain information about their internal parameters (privacy breaches). The main contributions of this chapter are twofold. First, we introduce a formal general framework for the empirical evaluation of the security of machine-learning systems. Second, according to our framework, we demonstrate the feasibility of evasion, poisoning and privacy attacks against SVMs in real-world security problems. For each attack technique, we evaluate its impact and discuss whether (and how) it can be countered through an adversary-aware design of SVMs. %IC
Our experiments are easily reproducible thanks to  \emph{open-source} code that we have made available, together with all the employed datasets, on a public repository.
}

\section{Introduction} 
\label{sect:introduction}

Machine-learning and pattern-recognition techniques are increasingly being adopted in security applications like spam filtering, network intrusion detection, and malware detection due to their ability to generalize, and to potentially detect novel attacks or variants of known ones.
Support Vector Machines (SVMs) are among the most successful techniques that have been applied for this purpose~\cite{drucker99,perdisci-ICDM06}. %add citations here related to the use of SVMs in security tasks

However, learning algorithms like SVMs assume \emph{stationarity}: that is, both the data used to train the classifier and the operational data it classifies are sampled from the same (though possibly unknown) distribution.
Meanwhile, in adversarial settings such as the above mentioned ones, intelligent and adaptive adversaries may purposely manipulate data (violating \emph{stationarity}) to exploit existing vulnerabilities of learning algorithms, and to impair the entire system.
This raises several open issues, related to whether machine-learning techniques can be safely adopted in security-sensitive tasks, or if they must (and can) be re-designed for this purpose. In particular, the main open issues to be addressed include:
\begin{compactenum}
\item analyzing the vulnerabilities of learning algorithms;
\item evaluating their security by implementing the corresponding attacks; and
\item eventually, designing suitable countermeasures.
\end{compactenum}

These issues are currently addressed in the emerging research area of \emph{adversarial machine learning}, at the intersection between computer security and machine learning.
This field is receiving growing interest from the research community, as witnessed by an increasing number of recent events: the NIPS Workshop on ``Machine Learning in Adversarial Environments for Computer Security'' (2007) \cite{nips07-adv}; the subsequent Special Issue of the Machine Learning journal titled ``Machine Learning in Adversarial Environments'' (2010) \cite{laskov10-ed}; the 2010 UCLA IPAM workshop on ``Statistical and Learning-Theoretic Challenges in Data Privacy''; the ECML-PKDD Workshop on ``Privacy and Security issues in Data Mining and Machine Learning'' (2010)~\cite{psdml10}; five consecutive CCS Workshops on  ``Artificial Intelligence and Security'' (2008-2012)~\cite{AISEC1,AISEC2,AISEC3,AISEC4,AISEC5}, and the Dagstuhl Perspectives Workshop on ``Machine Learning for Computer Security'' (2012)~\cite{dagstuhl12-adv}.

In Section~\ref{sect:background}, we review the literature of
adversarial machine learning, focusing mainly on the issue of
\emph{security evaluation}. We discuss both theoretical work and
applications, including examples of how learning can be attacked in
practical scenarios, either during its training phase (\ie,
\emph{poisoning} attacks that contaminate the learner's training data
to mislead it) or during its deployment phase (\ie, \emph{evasion}
attacks that circumvent the learned classifier).

In Section~\ref{sect:framework}, we summarize our recently defined
framework for the empirical evaluation of classifiers'
security~\cite{biggio12-tkde}. It is based on a \emph{general} model
of an adversary that builds on previous models and guidelines proposed
in the literature of adversarial machine learning.  We expound on the
assumptions of the adversary's goal, knowledge and capabilities that
comprise this model, which also easily accommodate application-specific
constraints. Having detailed the assumptions of his adversary, a
security analyst can formalize the adversary's strategy as an
optimization problem.

We then demonstrate our framework by applying it to assess the
security of SVMs.
%BB: We devise evasion attacks against the SVM in
We discuss our recently devised evasion attacks against SVMs~\cite{biggio13-ecml} in
Section~\ref{sect:evasion},
and review and extend our recent work~\cite{biggio12-icml} on
poisoning attacks against SVMs in
Section~\ref{sect:poisoning}.  We show that the optimization problems
corresponding to the above attack strategies can be solved through
simple gradient-descent algorithms.  The experimental results for
these evasion and poisoning attacks show that the SVM is vulnerable to
these threats for both linear and non-linear kernels in several
realistic application domains including handwritten digit
classification and malware detection for PDF files. We further explore
the threat of privacy-breaching attacks aimed at the SVM's training data
in Section~\ref{sect:privacy} where we apply our
framework to precisely describe the setting and threat model.
%, and touch on the open problem of privacy attacks on SVMs.

Our analysis provides useful insights into the potential security
threats from the usage of learning algorithms (and, particularly, of
SVMs) in real-world applications, and sheds light on whether they can
be safely adopted for security-sensitive tasks.  The presented
analysis allows a system designer to \emph{quantify} the security
risk entailed by an SVM-based detector so that he may weigh it against the benefits provided by the learning.
% be conscious of the risks of operating these classifiers.
It further suggests guidelines and countermeasures that
may mitigate threats and thereby improve overall system
security. These aspects are discussed for evasion and poisoning
attacks in Sections \ref{sect:evasion} and~\ref{sect:poisoning}.  In
Section~\ref{sect:privacy} we focus on developing countermeasures for
privacy attacks that are endowed with strong theoretical guarantees
within the framework of \emph{differential privacy}. We
conclude with a summary and discussion in
Section~\ref{sect:conclusions}.

In order to support the reproducibility of our experiments, we published all the code and the data employed for the experimental evaluations described in this paper~\cite{advlib}. In particular, our code is released under open-source license, and carefully documented, with the aim of allowing other researchers to not only reproduce, but also customize, extend and improve our work.

%DM: my comments are always referred to the paragraph above them
%DM: exchanged sect. 4 e 5

\section{Background} 
\label{sect:background}

In this section, we review the main concepts used throughout this chapter. 
We first introduce our notation and summarize the SVM learning problem. We then motivate the need for the proper assessment of the security of a learning algorithm so that it can be applied to security-sensitive tasks.

Learning can be generally stated as a process by which data is used to form a hypothesis that performs better than an \emph{a priori} hypothesis formed without the data. For our purposes, the hypotheses will be represented as functions of the form  $f : \mathcal X \to \mathcal Y$, which assign an input sample point $\vec x \in \mathcal X$ to a class $y \in \mathcal Y$; that is, given an observation from the input space $\mathcal X$, a hypothesis $f$ makes a prediction in the output space $\mathcal Y$. For \emph{binary classification}, the output space is binary and we use $\mathcal Y = \{ -1, +1\}$. In the classical \emph{supervised learning} setting, we are given a paired training dataset $ \{ (\mathbf x_{i}, y_{i}) \; | \; \mathbf x_{i} \in \mathcal{X}, y_{i} \in \mathcal Y \}_{i=1}^{n}$, we assume each pair is drawn independently from an unknown joint distribution $P(\vec X,Y)$, and we want to infer a classifier $f$ able to \emph{generalize} well on $P(\vec X,Y)$; \ie, to accurately predict the label $y$ of an unseen sample $\vec x$ drawn from that distribution.

\subsection{Support Vector Machines}
\label{sect:background-svm}

In its simplest formulation, an SVM learns a linear classifier for a binary classification problem. Its decision function is thus
$f(\mathbf x) = \mbox{sign}(  \vec{w}^{\top}\vec{x} + b )$,
where $\mbox{sign}(a) = +1$ ($-1$) if $a\geq0$ ($a<0$), and $\vec w$ and $b$ are learned parameters that specify the position of the decision hyperplane in feature space: the hyperplane's normal $\vec w$ gives its orientation and $b$ is its displacement. The learning task is thus to find a hyperplane that well-separates the two classes.  While many hyperplanes may suffice for this task, the SVM hyperplane both separates the training samples of the two classes and provides a maximum distance from itself to the nearest training point (this distance is called the classifier's \emph{margin}), since maximum-margin learning generally reduces \emph{generalization error}~\cite{vapnik95-book}.
Although originally designed for linearly-separable classification tasks (\emph{hard-margin} SVMs), SVMs were extended to non-linearly-separable classification problems by Vapnik~\cite{vapnik95} (\emph{soft-margin} SVMs), which allow some samples to violate the margin. In particular, a soft-margin SVM is learned by solving the following convex quadratic program (QP):
\begin{equation*}
  \begin{aligned}
    & \underset{\vec w, b, \xi}{\text{min}}
    & & \frac{1}{2} \vec{w}^{\top}\vec{w}
    + C \sum_{i=1}^{n} \xi_{i} \\
    & \text{s. t.}
     & & \forall \; i=1,\ldots,n \quad y_i(\vec{w}^{\top}\vec{x}_i + b) \geq 1 - \xi_{i} \quad \textrm{and} \quad \xi_{i} \geq 0 
  \enspace,
  \end{aligned}
%\label{eq:primal}
\end{equation*}
where the margin is maximized by minimizing $\frac{1}{2} \vec{w}^{\top} \vec{w}$, and the variables $\xi_{i}$ (referred to as \emph{slack variables}) represent the extent to which the samples, $\vec x_{i}$, violate the margin.
The parameter $C$ tunes the trade-off between minimizing the sum of the slack violation errors and maximizing the margin.

While the primal can be optimized directly, it is often solved via its
(Lagrangian) dual problem written in terms of Lagrange multipliers,
$\alpha_{i}$, which are constrained so that
$\sum_{i=1}^n \alpha_i y_i = 0$ and $0 \leq \alpha_{i} \leq C$ for
$i=1,\ldots,n$. Solving the dual has a
computational complexity that grows according to the size of the
training data as opposed to the feature space's dimensionality.  Further,
in the dual formulation, both the data and the slack variables
become implicitly represented---the data is represented by a
\emph{kernel matrix}, $\vec K$, of all inner products between pairs of
data points (that is, $K_{i,j} = \vec{x}_i^{\top}\vec{x}_j$) and each
slack variable is associated with a Lagrangian
multiplier via the KKT conditions that arise from duality. Using the
method of Lagrangian multipliers, the \emph{dual problem} is derived,
in matrix form, as
\begin{equation*}
\begin{aligned}
& \underset{\alpha}{\text{min}}
& & \frac{1}{2} \alpha^{\top} \vec Q \alpha - \vec 1_{n}^{\top} \alpha \\
& \text{s. t.}
& & \sum_{i=1}^{n}  \alpha_{i} y_{i} = 0 \quad \textrm{and} \quad \forall \; i=1,\ldots,n \quad 0 \leq \alpha_{i} \leq C \enspace, \\
%& & & \sum_{i=1}^{n}  \alpha_{i} y_{i} = 0 ,
\end{aligned}
%\label{eq:dual-matrix}
\end{equation*}
where $\vec Q = \vec K \circ yy^{\top}$ (the Hadamard product of $\vec K$ and $yy^{\top}$) and $\vec 1_{n}$ is a vector of $n$ ones. 

Through the kernel matrix, SVMs can be extended to more complex feature spaces (where a linear classifier may perform better) via a \emph{kernel function}---an implicit inner product from the alternative feature space.
%The kernel representation of the data allows one to build linear classifiers in higher-dimensional feature spaces without first explicitly mapping the data into that space; 
That is, if some function $\phi : \mathcal X \to \Phi$ maps training samples into a higher-dimensional feature space, then $K_{ij}$ is computed via the space's corresponding kernel function, $\kappa(\vec{x}_{i},\vec{x}_{j}) = \phi( \vec{x}_{i})^{\top} \phi(\vec{x}_{j})$.
Thus, one need not explicitly know $\phi$, only its corresponding \emph{kernel function}.
%This is known as the \emph{kernel trick}.

Further, the dual problem and its KKT conditions elicit interesting properties of the SVM.
First, the optimal primal hyperplane's normal vector, $\vec w$, is a linear combination of the training samples;\footnote{This is an instance of the Representer Theorem which states that solutions to a large class of regularized ERM problems lie in the span of the training data~\cite{representer}.} \ie, $\vec w = \sum_{i=1}^n \alpha_i y_i \vec x_i$.
 Second, the dual solution is \emph{sparse}, and only samples that lie on or within the hyperplane's margin have a  non-zero $\alpha$-value. 
Thus, if $\alpha_{i} = 0$, the corresponding sample $\vec x_{i}$ is correctly classified, lies beyond the margin (\ie, $y_{i}(\vec{w}^{\top}\vec{x}_{i} + b)>1$) and is called a \emph{non-support vector}.
If $\alpha_{i} = C$, the \nth{i}{th} sample violates the margin (\ie, $y_{i}(\vec{w}^{\top} \vec{x}_{i} + b) < 1$) and is an \emph{error vector}. Finally, if $0 < \alpha_{i} < C$, the \nth{i}{th} sample lies exactly on the margin (\ie, $y_{i}(\vec{w}^{\top}\vec{x}_{i} + b)=1$) and is a \emph{support vector}. As a consequence, the optimal displacement $b$ can be determined by averaging $y_{i} - \vec{w}^{\top}\vec{x}_{i}$ over the support vectors.

\subsection{Machine Learning for Computer Security: Motivation, Trends, and Arms Races}
\label{sect:ml-for-security}

In this section, we motivate the recent adoption of machine-learning techniques in computer security and discuss the novel issues this trend raises.
In the last decade, security systems increased in complexity to counter the growing sophistication and variability of attacks; a result of a long-lasting and continuing arms race in security-related applications such as malware detection, intrusion detection and spam filtering. The main characteristics of this struggle and the typical approaches pursued in security to face it are discussed in Section~\ref{sect:arms-race}. 
We now discuss some examples that better explain this trend and motivate the use of modern machine-learning techniques for security applications.

In the early years, the attack surface (\ie, the vulnerable points of a system) of most systems was relatively small and most attacks were simple. In this era, signature-based detection systems (\eg, rule-based systems based on string-match\-ing techniques) were considered sufficient to provide an acceptable level of security. However, as the complexity and exposure of sensitive systems increased in the Internet Age, more targets emerged and the incentive for attacking them became increasingly attractive, thus providing a means and motivation for developing sophisticated and diverse attacks. 
Since signature-based detection systems can only detect attacks matching an existing \emph{signature}, attackers used minor variations of their attacks to evade detection (\eg, string-matching techniques can be evaded by slightly changing the attack code).
To cope with the increasing variability of attack samples and to
detect never-before-seen attacks, machine-learning approaches have
been increasingly incorporated into these detection systems
to complement traditional signature-based detection.
These two approaches can be combined to make accurate and agile
detection: signature-based detection offers fast and lightweight
filtering of most known attacks, while machine-learning approaches can
process the remaining (unfiltered) samples and identify new (or less
well-known) attacks.
%DM: corrected:  "since signature-based detection...."
%DM: Added sentence after "evade detection"
%DM: "machine learning approaches have been"
%DM: removed "and not merely substituting"
%DM: machine learning approaches can be used..

\textbf{The quest of image spam}. A recent example of the above arms race is \emph{image spam} (see, \eg, \cite{biggio-PRL11}). In 2006, to evade the  textual-based spam filters, spammers began rendering their messages into images included as attachments, thus producing ``image-based spam,'' or \emph{image spam} for short. Due to the massive volume of image spam sent in 2006 and 2007, researchers and spam-filter designers proposed several different countermeasures. Initially, suspect images were analyzed by OCR tools to extract text for standard spam detection, and then signatures were generated to block the (known) spam images. However, spammers immediately reacted by \emph{randomly} obfuscating images with \emph{adversarial} noise, both to make OCR-based detection ineffective, and to evade signature-based detection. The research community responded with (fast) approaches mainly based on machine-learning techniques using visual features extracted from images, which could accurately discriminate between spam images and legitimate ones (\eg, photographs, plots, \etc).
Although image spam volumes have since declined, the exact cause for this decrease is debatable---these countermeasures may have played a role, but the image spam were also more costly to the spammer as they required more time to generate and more bandwidth to deliver, thus limiting the spammers' ability to send a high volume of messages. Nevertheless, had this arms race continued, spammers could have attempted to evade the countermeasures by \emph{mimicking} the feature values exhibited by legitimate images, which would have, in fact, forced spammers to increase the number of colors and elements in their spam images thus further increasing the size of such files, and the cost of sending them.

%DM:  machine learning techniques
%DM / BB: added "and the cost of sending them" to "the size"
\textbf{Misuse and anomaly detection in computer networks}. Another example of the above arms race can be found in network intrusion detection, where \emph{misuse detection} has been gradually augmented by \emph{anomaly detection}. The former approach relies on detecting attacks on the basis of signatures extracted from (known) intrusive network traffic, while the latter is based upon a statistical model of the \emph{normal profile} of the network traffic and detects \emph{anomalous} traffic that deviates from the assumed model of normality. This model is often constructed using machine-learning techniques, such as one-class classifiers (\eg, one-class SVMs), or, more generally, using density estimators.
The underlying assumption of anomaly-detection-based intrusion detection, though, is that \emph{all} \emph{anomalous} network traffic is, in fact, intrusive. Although intrusive traffic often does exhibit anomalous behavior, the opposite is not necessarily true: some non-intrusive network traffic may also behave anomalously. Thus, accurate anomaly detectors often suffer from high false-alarm rates.
%DM: removed "in recent years"

\subsection{Adversarial Machine Learning}
\label{sect:adv-ml}

As witnessed by the above examples, the introduction of
machine-learning techniques in security-sensitive tasks has many
beneficial aspects, and it has been somewhat necessitated by the
increased sophistication and variability of recent attacks and
zero-day exploits.  However, there is good reason to believe that
machine-learning techniques themselves will be subject to carefully
designed attacks in the near future, as a logical next step in the
above-sketched arms race.  Since machine-learning techniques were not
originally designed to withstand manipulations made by intelligent and
adaptive adversaries, it would be reckless to naively trust these
learners in a secure system.  Instead, one needs
to carefully consider whether these techniques can introduce novel
vulnerabilities that may degrade the overall system's security, or
whether they can be safely adopted.  In other words, we need to
address the question raised by Barreno \etal~\cite{barreno-ASIACCS06}:
\emph{can machine learning be secure?}

At the center of this question is the effect an adversary can have on
a learner by violating the \emph{stationarity assumption} that the
\emph{training data} used to train the classifier comes from the same
distribution as the \emph{test data} that will be classified by the
learned classifier.  This is a conventional and natural assumption
underlying much of machine learning and is the basis for
performance-evaluation-based techniques like cross-validation and
bootstrapping as well as for principles like empirical risk
minimization (ERM).  However, in security-sensitive settings, the adversary
may purposely manipulate data to mislead learning.  Accordingly, the
data distribution is subject to \emph{change}, thereby potentially
violating non-stationarity, albeit, in a limited way subject to the adversary's assumed capabilities (as we discuss in Section~\ref{sect:adversary-capability}).  Further, as in
most security tasks, predicting how the data
distribution will change is difficult, if not impossible
\cite{biggio12-tkde,huang11}.  Hence, adversarial learning problems
are often addressed as a \emph{proactive} arms race
\cite{biggio12-tkde}, in which the classifier designer tries to
anticipate the next adversary's move, by simulating and hypothesizing
proper attack scenarios, as discussed in the next section.

\subsubsection{Reactive and Proactive Arms Races}
\label{sect:arms-race}

\begin{figure}[htbp]
\begin{center}
\begin{minipage}{0.75\textwidth}
  \includegraphics[width=\textwidth]{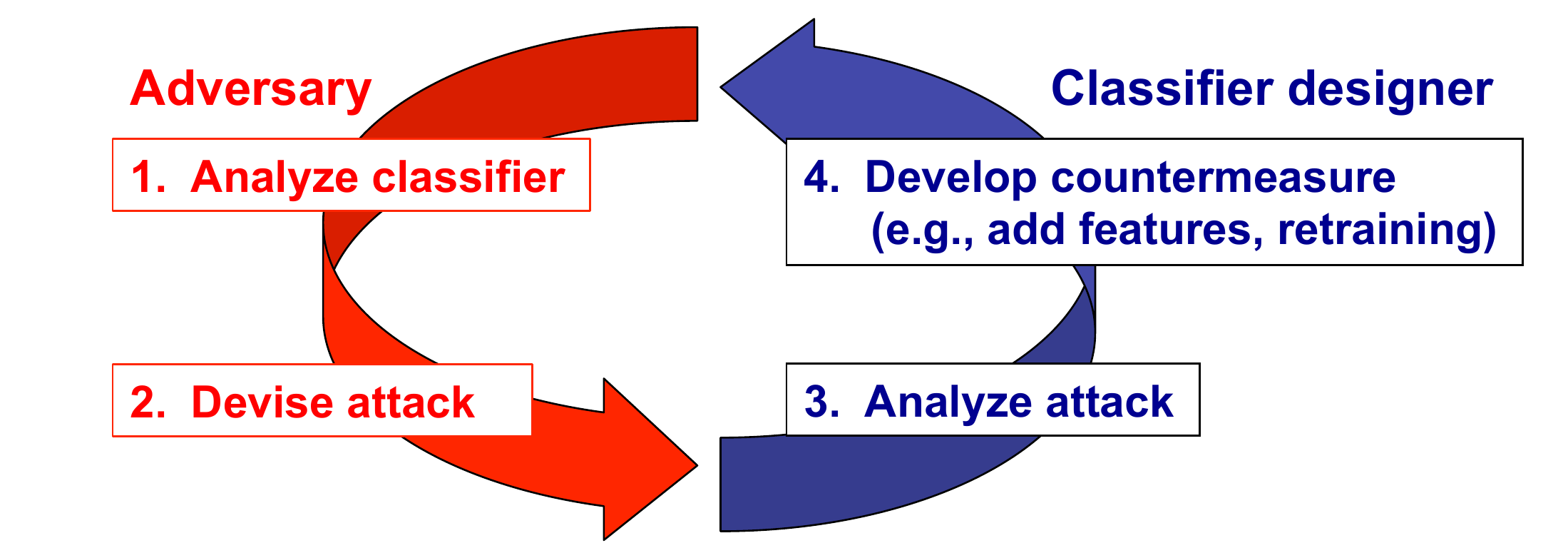}
  \caption{A conceptual representation of the \emph{reactive} arms race \cite{biggio12-tkde}.}
  \label{fig:arms-race-reactive}
\end{minipage}
\end{center}
\end{figure}

As mentioned in the previous sections, and highlighted by the examples
in Section~\ref{sect:ml-for-security}, security problems are often
cast as a long-lasting \emph{reactive} arms race between the
classifier designer and the adversary, in which each player attempts
to achieve his/her goal by reacting to the changing behavior of
his/her opponent. For instance, the adversary typically crafts
samples to evade detection (\eg, a spammer's goal
is often to create spam emails that will not be detected), while the
classifier designer seeks to develop a system that accurately
detects most malicious samples while maintaining a very low
false-alarm rate; \ie, by not falsely identifying legitimate examples.
Under this setting, the arms race can be modeled as the following
cycle~\cite{biggio12-tkde}.  First, the adversary analyzes the
existing learning algorithm and manipulates her data to evade
detection (or more generally, to make the learning algorithm
ineffective). For instance, a spammer may gather some knowledge of the
words used by the targeted spam filter to block spam and then
manipulate the textual content of her spam emails accordingly; \eg, words
like ``cheap'' that are indicative of spam can be
misspelled as ``che4p''.  Second, the classifier designer reacts by
analyzing the novel attack samples and updating his classifier.  This
is typically done by retraining the classifier on the newly collected
samples, and/or by adding features that can better detect the novel
attacks.  In the previous spam example, this amounts to retraining
the filter on the newly collected spam and, thus, to adding novel
words into the filter's dictionary (\eg, ``che4p'' may be now learned
as a spammy word). This \emph{reactive} arms race continues in
perpetuity as illustrated in Figure~\ref{fig:arms-race-reactive}.

\begin{figure}[htbp]
\begin{center}
\begin{minipage}{0.75\textwidth}
  \includegraphics[width=\textwidth]{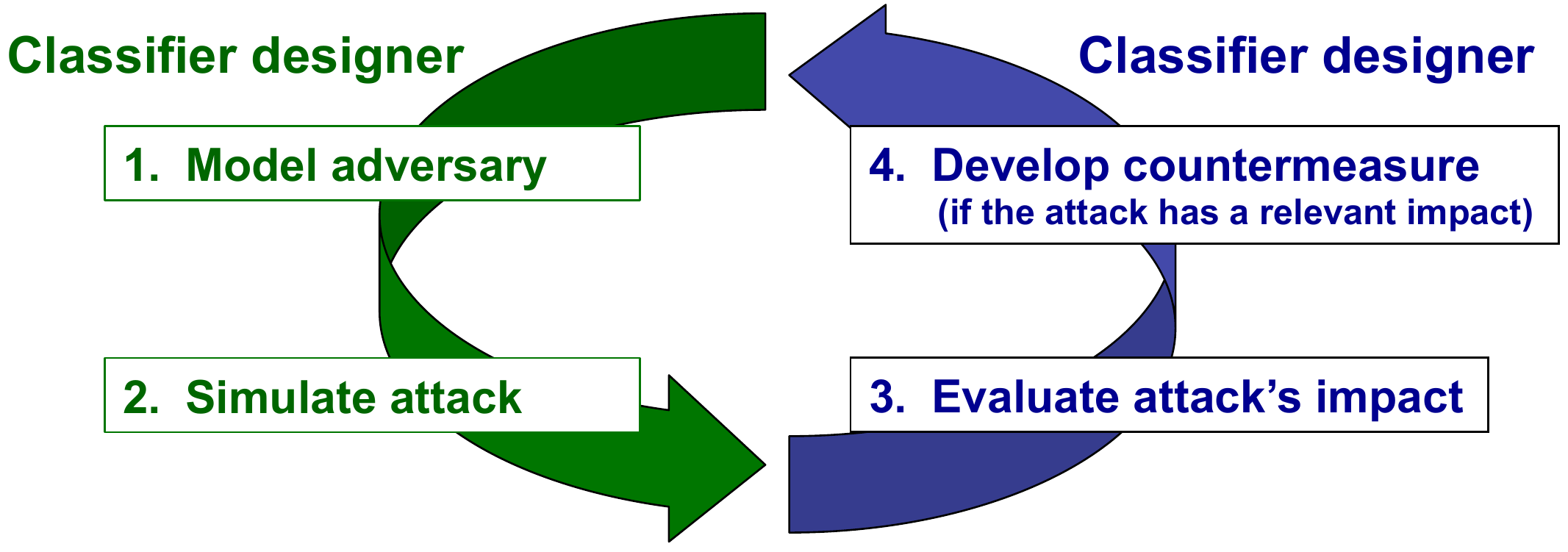}
  \caption{A conceptual representation of the \emph{proactive} arms race \cite{biggio12-tkde}.}
  \label{fig:arms-race-proactive}
\end{minipage}
\end{center}
\end{figure}

However, \emph{reactive} approaches to this arms race do not anticipate the next generation of security vulnerabilities and thus, the system potentially remains vulnerable to new attacks.
Instead, computer security guidelines traditionally advocate a \emph{proactive} approach\footnote{Although in certain abstract models we have shown how regret-minimizing online learning can be used to define reactive approaches that are competitive with proactive security~\cite{reactive}.}---the classifier designer should \emph{proactively anticipate} the adversary's strategy by (i) identifying the most relevant threats, (ii) designing proper countermeasures into his classifier, and (iii) repeating this process for his new design \emph{before} deploying the classifier. 
This can be accomplished by modeling the adversary (based on knowledge of the adversary's goals and capabilities) and using this model to simulate attacks, as is depicted in Figure~\ref{fig:arms-race-proactive} to contrast the reactive arms race.
While such an approach does not account for unknown or changing aspects of the adversary, it can indeed lead to an improved level of security by delaying each step of the \emph{reactive} arms race because it should reasonably force the adversary to exert greater effort (in terms of time, skills, and resources) to find new vulnerabilities.
Accordingly, proactively designed classifiers should remain useful for a longer time, with less frequent supervision or human intervention and with less severe vulnerabilities.

Although this approach has been implicitly followed in most of the previous work (see Section~\ref{sect:background-adversarial}), it has only recently been formalized within a more general framework for the empirical evaluation of a classifier's security \cite{biggio12-tkde}, which we summarize in Section~\ref{sect:framework}. 
Finally, although security evaluation may suggest specific countermeasures,
%the design of general-purpose \emph{secure} classifiers remains a distinct open problem.
designing general-purpose \emph{secure} classifiers remains an open problem.

\subsubsection{Previous Work on Security Evaluation}
\label{sect:background-adversarial}

%WE SHOULD ADD HERE MORE DESCRIPTION ON PRIVACY AND DESIGN OF COUNTERMEASURES

%\textbf{BAN: This is a repetition from the Introduction. We can remove this paragraph from here (probably merge it with text in the Introduction).}

%To study the effects of adversarial non-stationarity on the security of learning, a new research community has been drawn together to address a multitude of issues at this intersection between machine learning and security, called \emph{adversarial} (or \emph{secure}) \emph{machine learning}.
%Within this discipline, there have been a number of recent workshops ``Machine Learning in Adversarial Environments''~\cite{nips07-adv}, ``Privacy and Security issues in Data Mining and Machine Learning''~\cite{psdml10} and ``Machine Learning Methods for Computer Security (Dagstuhl Perspectives Workshop 12371)''~\cite{dagstuhl12-adv}; an ongoing series of workshops entitled the ``Workshop on Artificial Intelligence and Security'' held from 2008--2012; a special issue of the Journal of Machine Learning~\cite{laskov10-ed}; and many papers and articles at major machine learning and security venues.

Previous work in adversarial learning can be categorized
according to the two main steps of the proactive arms race described
in the previous section. The first research direction focuses on
identifying potential vulnerabilities of learning algorithms and
assessing the impact of the corresponding attacks on the targeted
classifier; \eg,
\cite{barreno10,barreno-ASIACCS06,cardenas-ws06,huang11,kloft12b,kolcz09,laskov09,lowd05}.
The second explores the development of proper
countermeasures and learning algorithms robust to known
attacks; \eg, \cite{dalvi04,kolcz09,rodrigues09}.

%In this chapter we focus on the evaluation of classifier security, that is, issue (1) above.
Although some prior work does address aspects of the
\emph{empirical} evaluation of classifier security, which is often implicitly
defined as the performance degradation incurred under a (simulated)
attack,
%I added an informal definition of classifier security here, ``en passant''
to our knowledge a systematic treatment of this process under a unifying perspective was only first described in our recent work \cite{biggio12-tkde}.
Previously, security evaluation is generally conducted within a
specific application domain such as spam filtering and network
intrusion detection (\eg, in
\cite{dalvi04,fogla06,kolcz09,lowd05-ceas,wittel04}), in which a
different application-dependent criteria is separately defined for
each endeavor.
Security evaluation is then implicitly undertaken by
defining an attack and assessing its impact on the given
classifier.
For instance, in \cite{fogla06}, the authors showed how
\emph{camouflage} network packets can mimic legitimate traffic to
evade detection; and, similarly, in
\cite{dalvi04,kolcz09,lowd05-ceas,wittel04}, the content of spam emails
was manipulated for evasion.
Although such analyses provide indispensable insights into
specific problems, their results are difficult to generalize to other
domains and provide little guidance for evaluating classifier security
in a different application.
Thus, in a new application domain, security evaluation often must
begin anew and it is difficult to directly compare with prior studies.
This shortcoming highlights the need for a more general set of
security guidelines and a more systematic definition of classifier
security evaluation, that we began to address in \cite{biggio12-tkde}.

Apart from application-specific work, several theoretical models of adversarial learning have been proposed~\cite{barreno10,brueckner12,dalvi04,huang11,kloft12b,laskov09,lowd05,nelson12-jmlr}. These models frame the secure learning problem and provide a foundation for a proper security evaluation scheme.
%but do not directly provide practical guidelines or tools for designing secure learning systems.
In particular, we build upon elements of the models of \cite{barreno10,barreno-ASIACCS06,huang11,kloft10,kloft12b,laskov09}, which were used in defining our framework for security evaluation \cite{biggio12-tkde}. Below we summarize these foundations.
%\textbf{BAN: is better NOW? now it should be clear that we have to recap the taxonomy and Pavel's work.}

\subsubsection{A Taxonomy of Potential Attacks against Machine Learning Algorithms} 
\label{sect:background-taxonomy}

%INSERT A FIGURE HERE HIGHLIGHTING THAT THE ADVERSARY CAN ALTER THE DATA (see huang11)

A taxonomy of potential attacks against pattern classifiers was proposed in~\cite{barreno10,barreno-ASIACCS06,huang11} as a baseline to characterize attacks on learners. 
The taxonomy is based on three main features: the kind of \emph{influence} of attacks on the classifier,  the kind of \emph{security violation} they cause, and the \emph{specificity} of an attack.
The attack's influence can be either {\bf causative}, if it aims to undermine learning, or {\bf exploratory}, if it targets the classification phase.
%Causative attacks result in producing at classification phase either many false-positive errors, making the classifier unusable, or false-negative errors, to cause malicious samples to be undetected. 
%Exploratory attacks, instead, may only produce false negative errors.
%Battista. The two sentences above are wrong, I think. Causative/Exploratory refers to manipulating training/testing and not on the kind of errors (this is instead the security violation)... I thus added the following sentence:
Accordingly, a causative attack may manipulate both training and testing data, whereas an exploratory attack only affects testing data. Examples of causative attacks include work in \cite{biggio12-icml,kloft10,kloft12b,nelson08,rubinstein09}, while exploratory attacks can be found in \cite{dalvi04,fogla06,kolcz09,lowd05-ceas,wittel04}.
The security violation can be either an {\bf integrity} violation, if it aims 
%to gain (non-confidential) knowledge of the classifier (\eg\ the learned classifiers' parameters) or
to gain unauthorized access to the system (\ie, to have malicious samples be misclassified as legitimate); an {\bf availability} violation, if the goal is to generate a high number of errors (both false-negatives and false-positives) such that normal system operation is compromised (\eg, legitimate users are denied access to their resources); or a {\bf privacy} violation, if it allows the adversary to obtain confidential information from the classifier (\eg, in biometric recognition, this may amount to recovering a protected biometric template of a system's client).
Finally, the attack specificity refers to the samples that are affected by the attack. It ranges continuously from {\bf targeted} attacks (\eg, if the goal of the attack is to have a specific spam email misclassified as legitimate) to {\bf indiscriminate} attacks (\eg, if the goal is to have \emph{any} spam email misclassified as legitimate).

Each portion of the taxonomy specifies a different type of attack as
laid out in Barreno~\etal~\cite{barreno10} and here we outline these
with respect to a PDF malware detector.  An example of a {\bf
  causative integrity} attack is an attacker who wants to mislead the
malware detector to falsely classify malicious PDFs as benign. The
attacker could accomplish this goal by introducing benign PDFs with
malicious features into the training set and the attack would be {\bf
  targeted} if the features corresponded to a particular malware or
otherwise an {\bf indiscriminate} attack. Similarly, the attacker
could cause a {\bf causative availability} attack by injecting malware
training examples that exhibited features common to benign messages;
again, these would be {\bf targeted} if the attacker wanted a
particular set of benign PDFs to be misclassified.  A {\bf causative
  privacy} attack, however, would require both manipulation of the
training and information obtained from the learned classifier. The
attacker could inject malicious PDFs with features identifying a
particular author and then subsequently test if other PDFs with those
features were labeled as malicious; this observed behavior may leak
private information about the authors of other PDFs in the training
set.

In contrast to the causative attacks, {\bf exploratory} attacks cannot
manipulate the learner, but can still exploit the learning mechanism. An example of
an {\bf exploratory integrity} attack involves an attacker who crafts
a malicious PDF for an existing malware detector. This attacker
queries the detector with candidate PDFs to discover which attributes
the detector uses to identify malware, thus, allowing her to re-design
her PDF to avoid the detector. This example could be {\bf targeted} to
a single PDF exploit or {\bf indiscriminate} if a set of possible exploits are
considered. An {\bf exploratory privacy} attack against the malware
detector can be conducted in the same way as the {\bf causative
  privacy} attack described above, but without first injecting PDFs
into the training data. Simply by probing the malware detector with
crafted PDFs, the attacker may divulge secrets from the detector.
Finally, {\bf exploratory availability} attacks are possible in some
applications but are not currently considered to be of interest.

\section{A Framework for Security Evaluation} 
\label{sect:framework}

In Sections~\ref{sect:adv-ml} and \ref{sect:arms-race}, we motivated the need for simulating a proactive arms race as a means for improving system security. We further argued that evaluating a classifier's security properties through simulations of different, potential attack scenarios is a crucial step in this arms race for identifying the most relevant vulnerabilities and for suggesting how to potentially counter them. 
% BB I also revised the sentence below
% DM: added "properties" to "classifier's security"
Here, we summarize our recent work~\cite{biggio12-tkde} that proposes a new framework for designing proactive secure classifiers by addressing the shortcomings of the reactive security cycle raised above.
Namely, our approach allows one to empirically evaluate a classifier's security during its design phase by addressing the first three steps of the proactive arms race depicted in Figure~\ref{fig:arms-race-proactive}: (i) identifying potential attack scenarios, (ii) devising the corresponding attacks, and (iii) systematically evaluating their impact.
Although it may also suggest countermeasures to the hypothesized attacks, the final step of the proactive arms race remains unspecified as a unique design step that has to be addressed separately in an application-specific manner.

Under our proposed security evaluation process, the analyst must clearly scrutinize the classifier by considering different attack scenarios to investigate a set of distinct potential vulnerabilities. This amounts to performing a more systematic \emph{what-if analysis} of classifier security~\cite{rizzi09}. This is an essential step in the design of security systems, as it not only allows the designer to identify the most important and relevant threats, but also it forces him/her to consciously decide whether the classifier can be reasonably deployed, after being made aware of the corresponding risks, or whether it is instead better to adopt additional countermeasure to mitigate the attack's impact \emph{before} deploying the classifier.

Our proposed framework builds on previous work and attempts to
systematize and unify their views under a more coherent perspective.
The framework defines how an analyst can conduct a security audit of a
classifier, which we detail in the remainder of this section.  First,
in Section~\ref{sect:adversary-model}, we explain how an adversary
model is constructed according to the adversary's anticipated goals,
knowledge and capabilities.  Based on this model, a simulation of the
adversary can be conducted to find the corresponding \emph{optimal}
attack strategies and produce simulated attacks, as described in
Section~\ref{sect:attack-strategy}.  These simulated attack samples
are then used to evaluate the classifier by either adding them to the
training or test data, in accordance with the adversary's capabilities
from Section~\ref{sect:adversary-capability}.  We conclude this
section by discussing how to exploit our framework in specific
application domains in Section~\ref{sect:howto}.

\subsection{Modeling the Adversary}
\label{sect:adversary-model}

The proposed model of the adversary is based on specific assumptions about her goal, knowledge of the system, and capability to modify the underlying data distribution by manipulating individual samples.
It allows the classifier designer to model the attacks identified in the attack taxonomy described as in Section~\ref{sect:background-taxonomy}~\cite{barreno10,barreno-ASIACCS06,huang11}. However, in our framework, one can also incorporate application-specific constraints into the definition of the adversary's capability. Therefore, it can be exploited to derive practical guidelines for developing optimal attack strategies and to guide the design of adversarially resilient classifiers.

\subsubsection{Adversary's Goal}
\label{sect:adversary-goal}

According to the taxonomy presented first by Barreno \etal~\cite{barreno-ASIACCS06} and extended by Huang \etal~\cite{huang11}, the adversary's goal should be defined based on the anticipated security violation, which might be an integrity, availability, or privacy violation (see Section~\ref{sect:background-taxonomy}), and also depending on the attack's specificity, which ranges from targeted to indiscriminate.
Further, as suggested by Laskov and Kloft~\cite{laskov09} and Kloft and Laskov~\cite{kloft12b}, the adversary's goal should be defined in terms of an objective function that the adversary is willing to maximize. This allows for a formal characterization of the \emph{optimal} attack strategy.

For instance, in an indiscriminate integrity attack, the adversary may aim to maximize the number of spam emails that evade detection, while minimally manipulating their content~\cite{dalvi04,lowd05,nelson12-jmlr}, whereas in an indiscriminate availability attack, the adversary may aim to maximize the number of classification errors, thereby causing a general denial-of-service due to an excess of false alarms~\cite{nelson08,biggio12-icml}.

\subsubsection{Adversary's Knowledge}
\label{sect:adversary-knowledge}

\begin{figure}[htbp]
\begin{center}
\begin{minipage}{0.9\textwidth}
  \includegraphics[width=\textwidth]{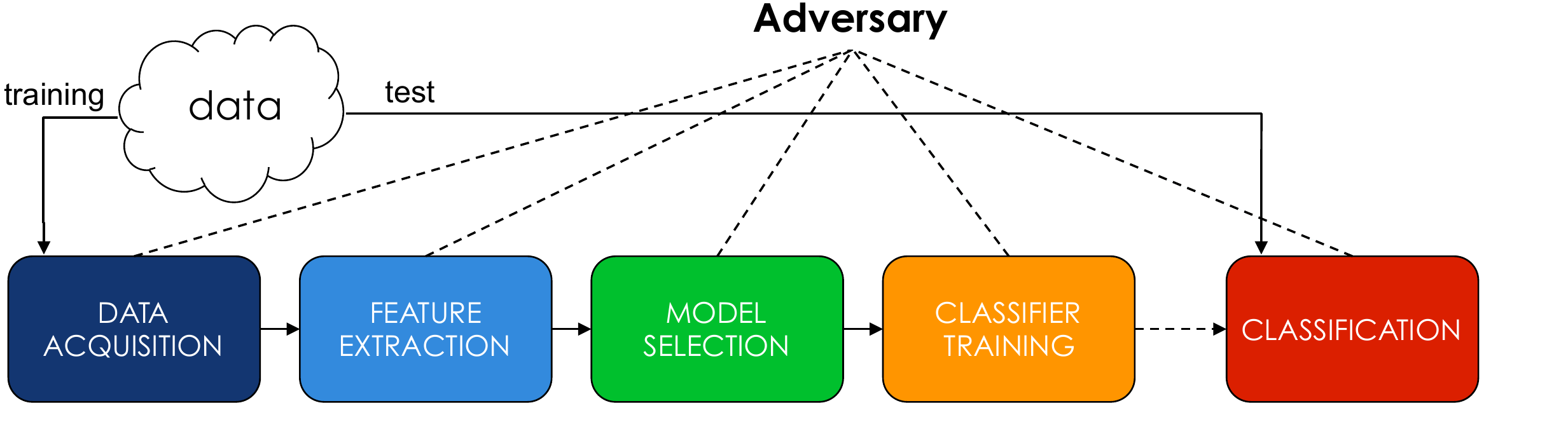}
  \caption{A representation of the design steps of a machine learning system \cite{duda-hart-stork} that may provide sources of knowledge for the adversary.}
  \label{fig:adversary-knowledge}
\end{minipage}
\end{center}
\end{figure}

The adversary's knowledge of the attacked system can be defined based on the main components involved in the design of a machine learning system, as described in~\cite{duda-hart-stork} and depicted in Figure~\ref{fig:adversary-knowledge}.

\begin{samepage}
According to the five design steps depicted in Figure~\ref{fig:adversary-knowledge}, the adversary may have various degrees of knowledge (ranging from no information to complete information) pertaining to the following five components:
\begin{compactenum}[(\itshape k.i\upshape)]
\item\label{know-1} the training set (or part of it);
\item\label{know-2} the feature representation of each sample; \ie, how \emph{real} objects (emails, network packets, \etc) are mapped into the feature space;
\item\label{know-3} the learning algorithm and its decision function; \eg, that logistic regression is used to learn a linear classifier;
\item\label{know-4} the learned classifier's parameters; \eg, the actual learned weights of a linear classifier; 
\item\label{know-5} feedback from the deployed classifier; \eg, the classification labels assigned to some of the samples by the targeted classifier.
\end{compactenum}
\end{samepage}

These five elements represent different levels of knowledge about the system being attacked.
A typical hypothesized scenario assumes that the adversary has \emph{perfect knowledge} of the targeted classifier \knowref{know-4}. Although potentially too pessimistic, this worst-case setting allows one to compute a lower bound on the classifier performance when it is under attack~\cite{dalvi04,kolcz09}.
A more realistic setting is that the adversary knows the (untrained) learning algorithm \knowref{know-3}, and she may exploit feedback from the classifier on the labels assigned to some \emph{query} samples \knowref{know-5}, either to directly find optimal or nearly-optimal attack instances~\cite{lowd05,nelson12-jmlr}, or to learn a surrogate classifier, which can then serve as a template to guide the attack against the actual classifier. We refer to this scenario as a \emph{limited knowledge} setting in Section~\ref{sect:evasion}.

Note that one may also make more restrictive assumptions on the adversary's knowledge, such as considering partial knowledge of the feature representation \knowref{know-2}, or a complete lack of knowledge of the learning algorithm \knowref{know-3}. Investigating classifier security against these uninformed adversaries may yield a higher level of security. However, such assumptions would be contingent on \emph{security through obscurity}; that is, the provided security would rely upon \emph{secrets} that must be kept unknown to the adversary even though such a high level of secrecy may not be practical. Reliance on unjustified secrets can potentially lead to catastrophic unforeseen vulnerabilities. Thus, this paradigm should be regarded as being complementary to \emph{security by design}, which instead advocates that systems should be designed from the ground-up to be secure and, if secrets are assumed, they must be well-justified. Accordingly, security is often investigated by assuming that the adversary knows at least the learning algorithm and the underlying feature representation.

\subsubsection{Adversary's Capability}
\label{sect:adversary-capability}

We now give some guidelines on how the attacker may be able to manipulate samples and the corresponding data distribution. As discussed in Section~\ref{sect:background-taxonomy}~\cite{barreno10,barreno-ASIACCS06,huang11}, the adversary may control both training and test data (causative attacks), or only on test data (exploratory attacks). Further, training and test data may follow different distributions, since they can be manipulated according to different attack strategies by the adversary. Therefore, we should specify:
\begin{compactenum}[(\itshape c.i\upshape)]
\item\label{cap-1} whether the adversary can manipulate training (TR) and/or testing (TS) data; \ie, the attack influence from the taxonomy in~\cite{barreno10,barreno-ASIACCS06,huang11});
\item\label{cap-2} whether and to what extent the attack affects the class priors, for TR and TS;
\item\label{cap-3} which and how many samples can be modified in each class, for TR and TS;
\item\label{cap-4} which features of each attack sample can be modified and how can these features' values be altered; \eg, correlated feature values can not be modified independently.
\end{compactenum}

%I would prefer not to give too many details on data distribution and resampling algorithm. I tried to condensate the information in these two paragraphs, but we should check consistency, and, most important, if this part is clear enough.

Assuming a generative model $p(\mathbf X,Y) = p(Y)p(\mathbf X |
Y)$ (where we use $p_{\rm tr}$ and $p_{\rm ts}$ for
training and test distributions, respectively), assumption
\capref{cap-2} specifies how an attack can modify the priors
$p_{\rm tr}(Y)$ and $p_{\rm ts}(Y)$ while assumptions \capref{cap-3}
and \capref{cap-4} specifies how it can alter the
class-conditional distributions $p_{\rm tr}(\mathbf X | Y)$ and
$p_{\rm ts}(\mathbf X | Y)$.

% If we assume the generative model $p(\mathbf X,Y) = p(Y)p(\mathbf X |
% Y)$ for training and test data, and respectively denote their
% distributions as $p_{\rm tr}(\mathbf X , Y)$ and $p_{\rm ts}(\mathbf X
% , Y)$, assumption \capref{cap-2} amounts to defining how the class
% priors $p_{\rm tr}(Y)$ and $p_{\rm ts}(Y)$ may be altered, while
% assumptions \capref{cap-3} and \capref{cap-4} implicitly define the
% class-conditional distributions $p_{\rm tr}(\mathbf X | Y)$ and
% $p_{\rm ts}(\mathbf X | Y)$.

To perform security evaluation according to the hypothesized attack scenario, it is thus clear that the collected data and generated attack samples should be resampled according to the above distributions to produce suitable training and test set pairs.
This can be accomplished through existing resampling algorithms like cross-validation or bootstrapping, when the attack samples are independently sampled from an identical distribution (i.i.d.). Otherwise, one may consider different sampling schemes. For instance, in Biggio \etal~\cite{biggio12-icml} the attack samples had to be injected into the training data, and each attack sample depended on the current training data, which also included past attack samples. In this case, it was sufficient to add one attack sample at a time, until the desired number of samples was reached.\footnote{See \cite{biggio12-tkde} for more details on the definition of the data distribution and the resampling algorithm.}

\subsubsection{Attack Strategy}
\label{sect:attack-strategy}

Once specific assumptions on the adversary's goal, knowledge, and capability are made, one can compute the optimal attack strategy corresponding to the hypothesized attack scenario; \ie, the adversary model. This amounts to solving the optimization problem defined according to the adversary's goal, under proper constraints defined in accordance with the adversary's assumed knowledge and capabilities.
The attack strategy can then be used to produce the desired attack samples, which then have to be merged consistently to the rest of the data to produce suitable training and test sets for the desired security evaluation, as explained in the previous section.
Specific examples of how to derive optimal attacks against SVMs, and how to resample training and test data to properly include them are discussed in Sections~\ref{sect:evasion} and~\ref{sect:poisoning}.

\subsection{How to use our Framework}
\label{sect:howto}

We summarize here the steps that can be followed to correctly use our framework in specific application scenarios: 
\begin{compactenum}
\item hypothesize an attack scenario by identifying a proper adversary's goal, and according to the taxonomy in \cite{barreno10,barreno-ASIACCS06,huang11};
\item define the adversary's knowledge according to \knowref[know-1]{know-5}, and capabilities according to \capref[cap-1]{cap-4};
\item formulate the corresponding optimization problem and devise the corresponding attack strategy;
\item resample the collected (training and test) data accordingly;
\item evaluate classifier's security on the resampled data (including attack samples);
\item repeat the evaluation for different levels of adversary's knowledge and/or capabilities, if necessary; or hypothesize a different attack scenario.
\end{compactenum}

In the next sections we show how our framework can be applied to
investigate three security threats to SVMs: evasion, poisoning,
and privacy violations. We then discuss how our findings may be used to improve
the security of such classifiers to the considered attacks.  For
instance, we show how careful kernel parameter selection, which
trades off between security to attacks and classification accuracy, may
complicate the adversary's task of subverting the learning process.

\section{Evasion Attacks against SVMs} %5-8 pages
\label{sect:evasion}

%Now we apply our framework and show how it can be exploited to devise optimal attacks against SVMs, with different goals. We consider here an evasion setting, where...

In this section, we consider the problem of SVM evasion at \emph{test time}; \ie, how to optimally manipulate samples at test time to avoid detection. The problem of evasion at test time has been considered in previous work, albeit either limited to simple decision functions such as linear classifiers~\cite{dalvi04,lowd05}, or to cover any convex-inducing classifiers~\cite{nelson12-jmlr} that partition the feature space into two sets, one of which is convex, but do not include most interesting families of non-linear classifiers such as neural nets or SVMs. In contrast to this prior work, the methods presented in our recent work~\cite{biggio13-ecml} %BB added citation to ecml
and in this section demonstrate that evasion of kernel-based classifiers at test time can be realized with a straightforward gradient-descent-based approach derived from Golland's technique of discriminative directions \cite{Gol02}. As a further simplification of the attacker's effort, we empirically show that, even if the adversary does not precisely know the classifier's decision function, she can learn a \emph{surrogate} classifier on a surrogate dataset and reliably evade the targeted classifier.

This section is structured as follows. In Section~\ref{sect:adv-model-evasion-test-time}, we define the model of the adversary, including her attack strategy, according to our evaluation framework described in Section~\ref{sect:adversary-model}. Then, in Section~\ref{subsect:evasion-algorithm} we derive the attack strategy that will be employed to experimentally evaluate evasion attacks against SVMs. We report our experimental results in Section~\ref{sect:exp-evasion-test-time}. Finally, we critically discuss and interpret our research findings in Section~\ref{subsect:discussion-evasion-test-time}.

\subsection{Modeling the Adversary}
\label{sect:adv-model-evasion-test-time}
\newcommand{\classifierLBL}[1][\vec x]{\ensuremath{f_{#1}}}

We show here how our framework can be applied to evaluate the
security of SVMs against evasion attacks.  We first introduce our notation,
state our assumptions about attack scenario, and then
derive the corresponding optimal attack strategy.

\textbf{Notation}. We consider a classification algorithm $f : \mathcal X \mapsto \mathcal Y$ that assigns samples represented in some feature space $\vec x \in \mathcal X$ to a label in the set of predefined classes $y \in \mathcal Y = \{-1,+1\}$, where $-1$ ($+1$) represents the legitimate (malicious) class. The label $\classifierLBL = f(\vec x)$ given by a classifier is typically obtained by thresholding a continuous discriminant function $g : \mathcal X \mapsto \R$. Without loss of generality, we assume that $f(\vec x) = -1$ if $g(\vec x) < 0$, and $+1$ otherwise. Further, note that we use $\classifierLBL$ to refer to a label assigned by the classifier for the point $\vec x$ (rather than the true label $y$ of that point) and the shorthand $\classifierLBL[i]$ for the label assigned to the \nth{i}{th} training point, $\vec x_i$.

\subsubsection{Adversary's Goal}
\label{subsect:adv-goal-evasion-test-time}

Malicious (positive) samples are manipulated to evade the classifier. The adversary may be satisfied when a sample $\vec x$ is found such that $g(\vec x) < -\epsilon$ where $\epsilon > 0$ is a small constant. However, as mentioned in Section \ref{sect:adversary-goal},  these attacks may be easily defeated by simply adjusting the decision threshold to a slightly more conservative value (\eg, to attain a lower false negative rate at the expense of a higher false positive rate). 
For this reason, we assume a \emph{smarter} adversary, whose goal is to have her attack sample misclassified as legitimate with the largest confidence. Analytically, this statement can be expressed as follows: find an attack sample $\vec x$ that minimizes the value of the classifier's discriminant function $g(\vec  x)$. Indeed, this adversarial setting provides a worst-case bound for the targeted classifier.

\subsubsection{Adversary's Knowledge}\label{subsect:adv-knowledge-evasion-test-time}
We investigate two adversarial settings.
%: \emph{perfect} knowledge (PK) and \emph{limited} knowledge (LK). 
In the first, the adversary has \emph{perfect} knowledge (PK) of the targeted classifier; \ie, she knows the feature space \knowref{know-2} and function $g(\vec x)$  \knowref[know-3]{know-4}. Thus, the labels from the targeted classifier \knowref{know-5} are not needed. In the second, the adversary is assumed to have \emph{limited} knowledge (LK) of the classifier. We assume she knows the feature representation \knowref{know-2} and the learning algorithm \knowref{know-3}, but that she does not know the learned classifier $g(\vec x)$ \knowref{know-4}. In both cases, we assume the attacker does not have knowledge of the training set \knowref{know-1}.

%BB: I rewrote some parts below.
%Blaine: So did I :)
Within the LK scenario, the adversary does not know the true
discriminant function $g(\vec x)$ but may approximate it as $\hat g(\vec x)$ by
learning a \emph{surrogate} classifier on a surrogate training set
$\{(\vec x_{i}, y_{i})\}_{i=1}^{n_{q}}$ of $n_{q}$ samples. This data may
be collected by the adversary in several ways; \eg, she may sniff
network traffic or collect legitimate and spam emails from an
alternate source. Thus, for LK, there are two sub-cases related to
assumption \knowref{know-5}, which depend on whether the adversary can
query the classifier.  If so, the adversary can build the training set
by submitting a set of $n_{q}$ queries $\vec x_{i}$ to the targeted
classifier to obtain their classification labels, $y_{i} = f(\vec x_{i})$.
This is indeed the adversary's true learning task, but it requires her
to have access to classifier feedback; \eg, by having an email account
protected by the targeted filter (for public email providers, the
adversary can reasonably obtain such accounts).  If not, the adversary
may use the true class labels for the surrogate data, although
this may not correctly approximate the targeted classifier (unless it is very
accurate).

\subsubsection{Adversary's Capability}
\label{subsect:adv-capability-evasion-test-time}

In the evasion setting, the adversary can only manipulate testing data \capref{cap-1}; \ie, she has no way to influence training data. We further assume here that the class priors can not be modified \capref{cap-2}, and that all the malicious testing samples are affected by the attack \capref{cap-3}. In other words, we are interested in simulating an \emph{exploratory}, \emph{indiscriminate} attack. The adversary's capability of manipulating the features of each sample \capref{cap-4} should be defined based on application-specific constraints. However, at a more general level we can bound the attack point to lie within some maximum distance from the original attack sample, $d_{\rm max}$, which then is a parameter of our evaluation. Similarly to previous work, the definition of a suitable distance measure $d : \mathcal X \times \mathcal X \mapsto \R$ is left to the specific application domain \cite{dalvi04,lowd05,nelson12-jmlr}. Note indeed that this distance should reflect the adversary's effort or cost in manipulating samples, by considering factors that can limit the overall attack impact; \eg, the increase in the file size of a malicious PDF, since larger files will lower the infection rate due to increased transmission times. For spam filtering, distance is often given as the number of modified words in each spam \cite{dalvi04,lowd05,nelson08,nelson12-jmlr}, since it is assumed that highly modified spam messages are less effectively able to convey the spammer's message.

\subsubsection{Attack Strategy}
\label{sect:attack-strategy-evasion-test-time}

Under the attacker's model described in Sections~\ref{subsect:adv-goal-evasion-test-time},~\ref{subsect:adv-knowledge-evasion-test-time} and~\ref{subsect:adv-capability-evasion-test-time}, for any target malicious sample $\vec x^{0}$ (the adversary's true objective), an optimal attack strategy finds a sample $\vec x^{\ast}$ to minimize $g$ or its estimate $\hat g$, subject to a bound on its modification distance from $\vec x^{0}$:
\begin{equation*}
%\label{eq:evasion-obj}
  \begin{aligned}
    \vec x^{\ast} = \arg \min_{\vec x} \; & \hat g(\vec x)  
    & \rm{s.t.} \quad & d(\vec x,\vec x^{0}) \leq d_{\rm max} \enspace. 
  \end{aligned}
\end{equation*}
For several classifiers, minimizing $g(\vec x)$ is equivalent to maximizing the estimated posterior $p(\classifierLBL=-1 | \vec x)$; \eg, for neural networks, since they directly output a posterior estimate, and for SVMs, since their posterior can be estimated as a sigmoidal function of the distance of $\vec x$ to the SVM hyperplane~\cite{platt99}.
%the distance of $x$ to the hyperplane can be used in a sigmoidal function as an estimate of the posterior~\cite{platt99}.
%NOTE: we should also say that more/other constraints can be considered. For instance, feature addition can be expressed as: $x^{0} \leq x$

\begin{figure*}[htbp]
\begin{center}
\includegraphics[width=0.49\textwidth]{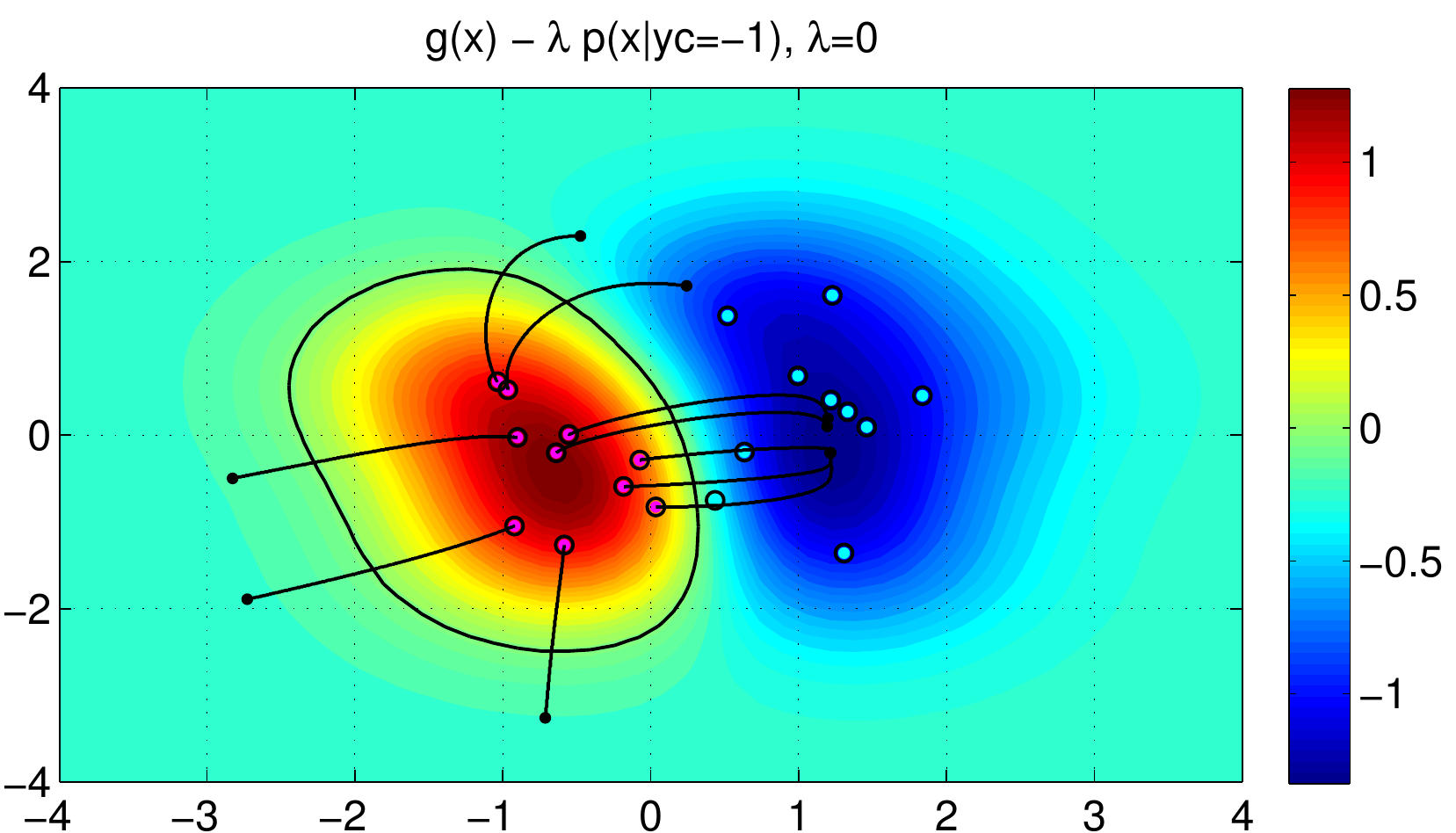}
\includegraphics[width=0.49\textwidth]{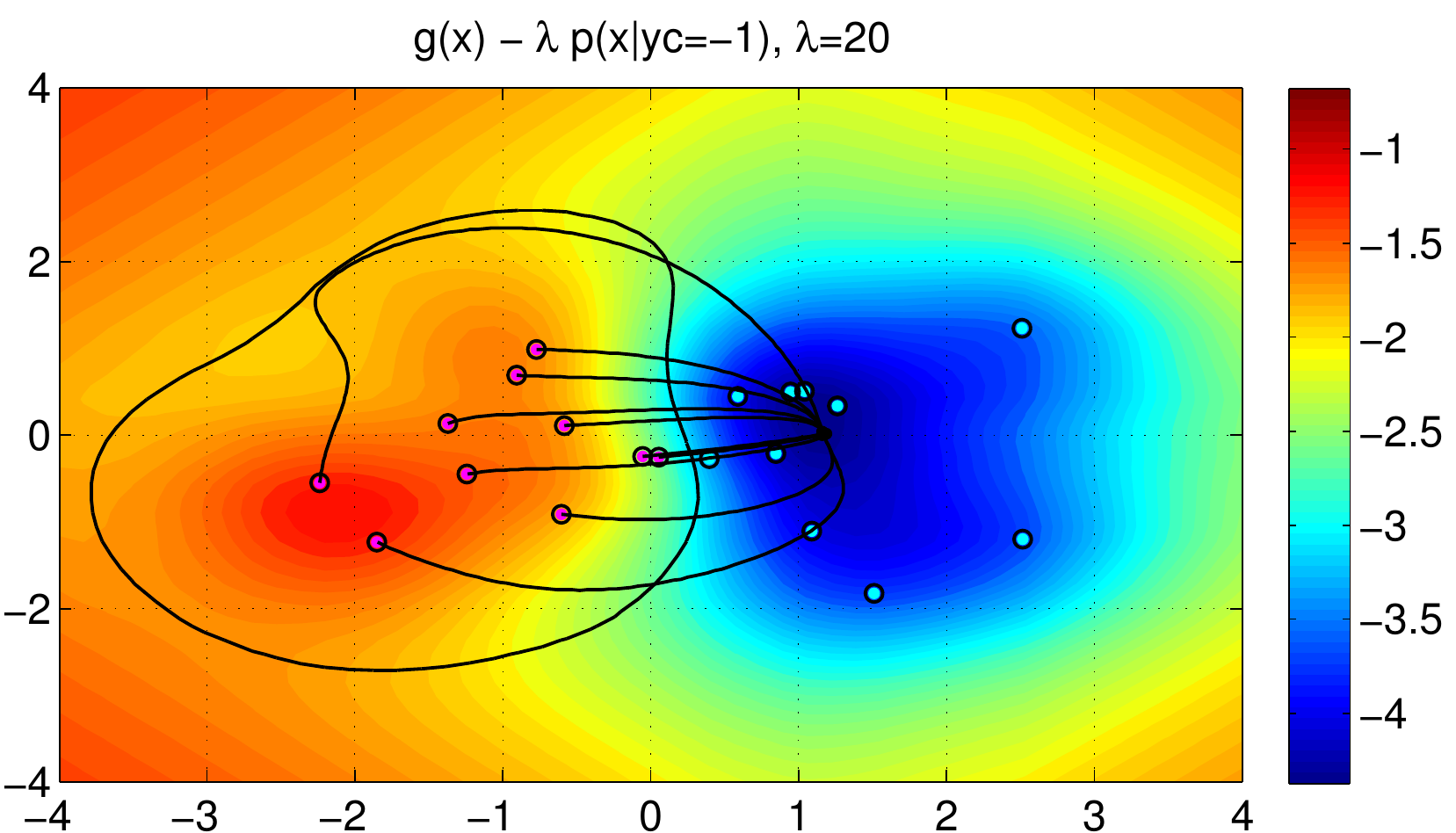}\\
\includegraphics[width=0.49\textwidth]{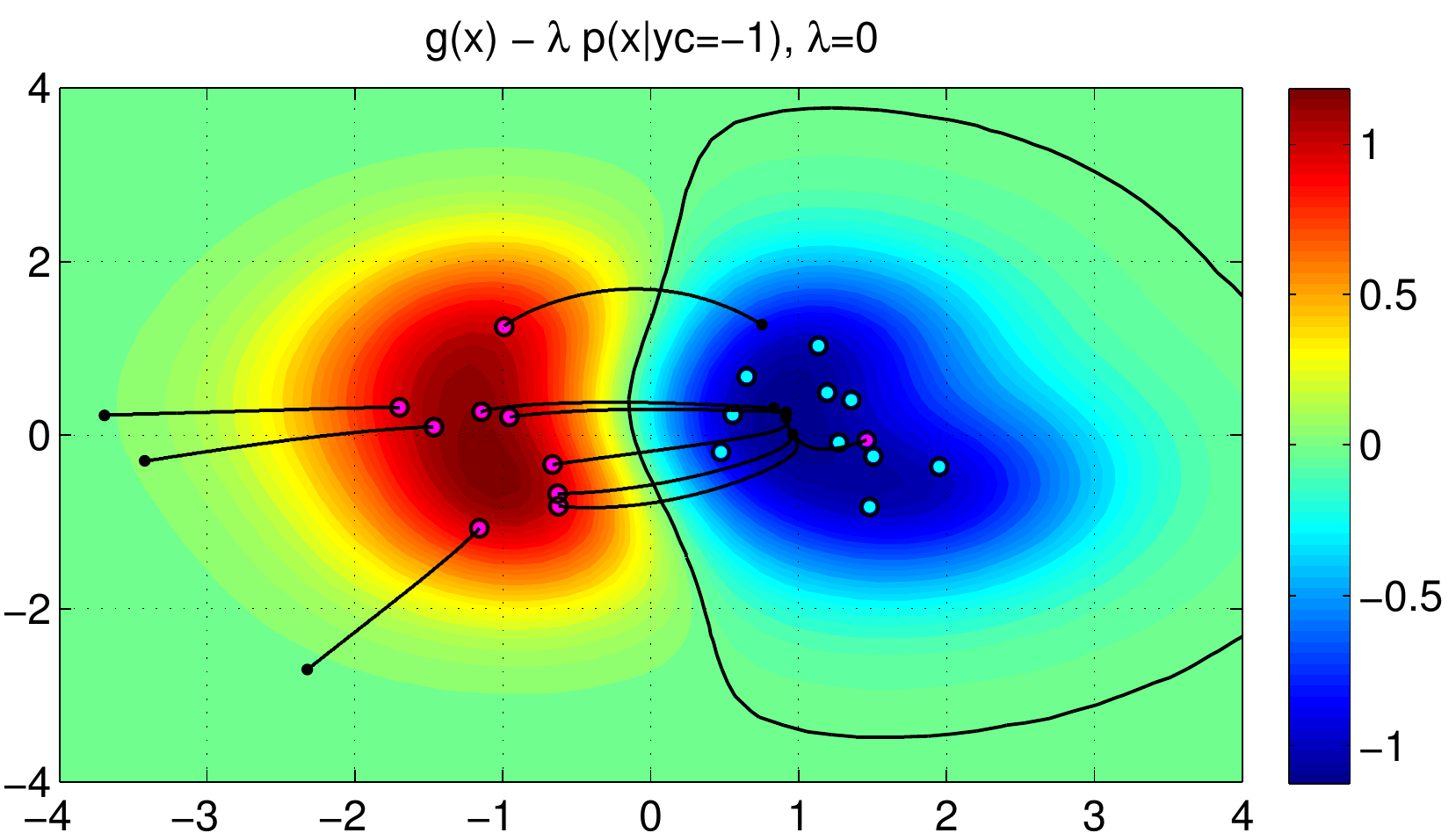}
\includegraphics[width=0.49\textwidth]{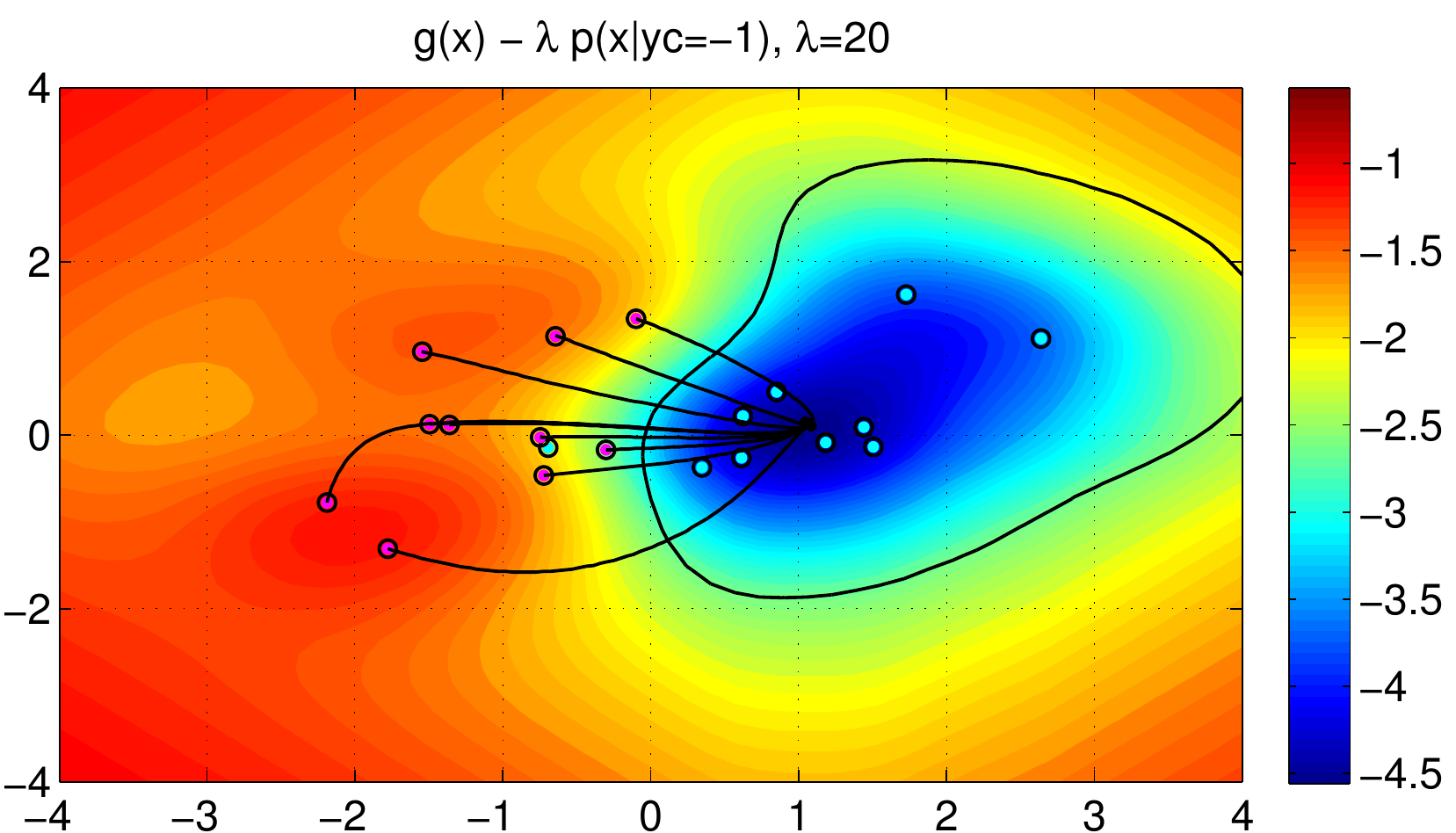}
\end{center}
\caption{Different scenarios for gradient-descent-based evasion procedures. In each, the function $g(\vec x)$ of the learned classifier is plotted with a color map with high values (red-orange-yellow) corresponding to the malicious class, low values (green-cyan-blue) corresponding to the benign class, and a black decision boundary for the classifier. For every malicious sample, we plot the path of a simple gradient descent evasion for a classifier with a closed boundary around the malicious class (\textbf{upper left}) or benign class (\textbf{bottom left}). Then, we plot the modified objective function of Equation~\eqref{eq:obj-function} and the paths of the resulting \algoName\ for a classifier with a closed boundary around the malicious (\textbf{upper right}) or benign class (\textbf{bottom right}).}
\label{fig:attack-strategy}
\end{figure*}

Generally, this is a non-linear optimization, which one may
optimize with many well-known techniques (\eg, gradient descent,
Newton's method, or BFGS) and below we use a gradient descent procedure.
However, if $\hat g(\vec x)$ is not convex, descent approaches may not find
a global optima.
%Thus, the optimum obtained may not achieve our evasion objective even though it may be otherwise possible. 
Instead, the descent path may lead to a flat region (local minimum) outside of the samples' support where $p(\vec x) \approx 0$ and the classification behavior of $g$ is unspecified and may stymie evasion attempts (see the \emph{upper left} plot in Figure~\ref{fig:attack-strategy}).

Unfortunately, our objective does not utilize the evidence we have about the distribution of data $p(\vec x)$, and thus gradient descent may meander into unsupported regions ($p(\vec x) \approx 0$) where $g$ is relatively unspecified. This problem is further compounded since our estimate $\hat{g}$ is based on a finite (and possibly small) training set making it a poor estimate of $g$ in unsupported regions, which may lead to false evasion points in these regions.
To overcome these limitations, we introduce an additional component into the formulation of our attack objective, which estimates $p(\vec x | \classifierLBL=-1)$ using density-estimation techniques. This second component acts as a penalizer for $\vec x$ in low density regions and is weighted by a parameter $\lambda \ge 0$ yielding the following modified optimization problem:
\begin{eqnarray}
\label{eq:constraint}
\label{eq:obj-function}
\arg \min_{\vec x} & &  E(\vec x) = \hat g(\vec x) - \frac{\lambda}{n} \sum_{i | \classifierLBL[i] = -1} k \left( \tfrac{\vec x-\vec x_{i}}{h} \right) \\
\nonumber \rm{s.t.} & & d(\vec x,\vec x^{0}) \leq d_{\rm max} \enspace ,
\end{eqnarray}
where $h$ is a bandwidth parameter for a kernel density estimator (KDE), and $n$ is the number of benign samples ($\classifierLBL=-1$) available to the adversary. %added the meaning of n here!
This alternate objective trades off between minimizing $\hat g(\vec x)$ (or $p(\classifierLBL=-1 | \vec x)$) and maximizing the estimated density $p(\vec x | \classifierLBL=-1)$. 
The extra component favors attack points to imitate features of known samples classified as legitimate, as in mimicry attacks~\cite{fogla06}. %maybe we can add a footnote here to better explain what a mimicry attack is?
In doing so, it reshapes the objective function and thereby biases the resulting \emph{\algoName} towards regions where the negative class is concentrated (see the \emph{bottom right} plot in Figure~\ref{fig:attack-strategy}). 

% Igino: added a few words to state that, on the other hand, the mimicry component alone may be not optimal, since it may completely fail to find ``blind spots''
Finally, note that this behavior may lead our technique to disregard attack patterns within unsupported regions ($p(\vec x) \approx 0$) for which $g(\vec x)<0$, when they do exist (see, \eg, the \emph{upper right} plot in Figure~\ref{fig:attack-strategy}). 
This may limit classifier evasion especially when the constraint $d(\vec x,\vec x^{0}) \leq d_{\rm max}$ is particularly strict. Therefore, the trade-off between the two components of the objective function should be carefully considered.

\subsection{Evasion Attack Algorithm}
\label{subsect:evasion-algorithm}%%

Algorithm~\ref{lab:attack} details a gradient-descent method for optimizing problem of Equation~\eqref{eq:obj-function}. %%BB: added ref to the optimization problem
It iteratively modifies the attack point $\vec x$ in the feature space as $\vec x^{\prime} \leftarrow \vec x- t \nabla E$, where $\nabla E$ is a unit vector aligned with the gradient of our objective function, and $t$ is the step size. We assume $g$ to be differentiable almost everywhere (subgradients may be used at discontinuities). When $g$ is non-differentiable or is not smooth enough for a gradient descent to work well, it is also possible to rely upon the mimicry / KDE term in the optimization of Equation~\eqref{eq:obj-function}. 

\begin{algorithm}[h]
  \caption{Gradient-descent attack procedure}
  \textbf{Input:} the initial attack point, $\vec x^{0}$; the step size, $t$; the trade-off parameter, $\lambda$; and $\epsilon > 0$. \\
  \textbf{Output:} $\vec x^{*}$, the final attack point.\\
  \begin{algorithmic}[1]
    \STATE{$k \gets 0$.}
    \REPEAT
          \STATE{$k \gets k + 1$}
      \STATE{Set $\nabla E(\vec x^{k-1})$ to a unit vector aligned
      with $\nabla g(\vec x^{k-1}) - \lambda \nabla{p(\vec x^{k-1} | \classifierLBL=-1)}$.} 
     \STATE{$\vec x^{k} \leftarrow \vec x^{k-1} - t \nabla E(\vec x^{k-1})$}
     \IF{$d(\vec x^{k},\vec x^{0}) > d_{\rm max}$}
     \STATE{Project $\vec x^{k}$ onto the boundary of the feasible region (enforcing application-specific constraints, if any).}
     \ENDIF
      \UNTIL{$E\left(\vec x^{k}\right) - E\left( \vec x^{k-1}\right) < \epsilon$}
      \STATE{\textbf{return:} $\vec x^{*} = \vec x^{k}$}
  \end{algorithmic}\label{lab:attack}
\end{algorithm}

In the next sections, we show how to compute the components of $\nabla E$; namely, the gradient of the discriminant function $g(\vec x)$ of SVMs for different kernels, and the gradient of the mimicking component (density estimation). We finally discuss how to project the gradient $\nabla E$ onto the feasible region in \emph{discrete} feature spaces.

\subsubsection{Gradient of Support Vector Machines}
\label{subsubsect:gradient-svm}

For SVMs, $g(\vec x) = \sum_i \alpha_i y_i k(\vec x,\vec x_{i}) + b$. The gradient is thus given by $\nabla g(\vec x) = \sum_i \alpha_i y_i \nabla k(\vec x,\vec x_{i})$. Accordingly, the feasibility of the approach depends on the computability of this kernel gradient $\nabla k(\vec x,\vec x_{i})$, which is computable for many numeric kernels. In the following, we report the kernel gradients for three main cases: (a) the linear kernel, (b) the RBF kernel, and (c) the polynomial kernel.

\textbf{(a) Linear kernel}. In this case, the kernel is simply given by $k(\vec x,\vec x_{i}) = \langle \vec x, \vec x_{i} \rangle$. Accordingly, $\nabla k(\vec x,\vec x_{i}) = \vec x_{i}$ (we remind the reader that the gradient has to be computed with respect to the current attack sample $\vec x$), and $\nabla g(\vec x) = \vec w = \sum_i \alpha_i y_i \vec x_{i}$.

\textbf{(b) RBF kernel}. For this kernel, $k(\vec x,\vec x_{i}) = \exp\{-\gamma \|\vec x-\vec x_{i}\|^2\}$. The gradient is thus given by $\nabla k(\vec x,\vec x_{i}) = -2 \gamma \exp\{-\gamma \|\vec x-\vec x_{i}\|^2\}(\vec x-\vec x_{i})$.

\textbf{(c) Polynomial kernel}. In this final case, $k(\vec x,\vec x_{i}) = (\langle \vec x, \vec x_{i} \rangle + c)^{p}$. The gradient is thus given by $\nabla k(\vec x,\vec x_{i}) = p(\langle \vec x, \vec x_{i} \rangle + c)^{p-1}\vec x_{i}$.

\subsubsection{Gradients of Kernel Density Estimators}
\label{sect:gradients-kde}

As with SVMs, the gradient of kernel density estimators depends on the gradient of its kernel.
We considered generalized RBF kernels of the form 
\begin{equation*}
  k \left (\tfrac{\vec x-\vec x_{i}}{h} \right ) = \exp{\left (  -\tfrac{d(\vec x,\vec x_{i})}{h}\right )},
\end{equation*}
where $d(\cdot,\cdot)$ is any suitable distance function. We used here the same distance $d(\cdot,\cdot)$ used in Equation~\eqref{eq:constraint},
but they can be different, in general.
For $\ell_{2}$- and $\ell_{1}$-norms (\ie, RBF and Laplacian kernels), the KDE (sub)gradients are respectively given by:
\begin{eqnarray*}
-\frac{2}{n h} \sum_{i | \classifierLBL[i] = -1}  \exp{\left (  -\frac{ \|\vec x-\vec x_{i}\|^{2}_2 }{h}\right )} (\vec x -\vec  x_{i}) \enspace, \\
-\frac{1}{n h} \sum_{i | \classifierLBL[i] = -1}  \exp{\left (  -\frac{ \|\vec x-\vec x_{i}\|_1}{h}\right )} (\vec x -\vec  x_{i}) \enspace.
\end{eqnarray*}

Note that the scaling factor here is proportional to $O(\frac{1}{nh})$.
Therefore, to influence gradient descent with a significant mimicking effect, the value of $\lambda$ in the objective function should be chosen such that the value of $\frac{\lambda}{nh}$ is comparable to (or higher than) the range of the discriminant function $\hat g(\vec x)$. 

\subsubsection{Gradient Descent Attack in Discrete Spaces}

In discrete spaces, gradient approaches may lead to a path through
infeasible portions of the feature space. In such cases, we need to find
feasible neighbors $\vec x$ that yield a steepest descent; \ie,
maximally decreasing $E(\vec x)$.  A simple approach to this problem is to
probe $E$ at every point in a small neighborhood of $\vec x$: $\vec x^\prime
\gets \arg\min_{\vec z \in \mathcal{N}(\vec x)}{E(\vec z)}$. However, this approach
requires a large number of queries. For classifiers
with a differentiable decision function, we can instead use the neighbor
whose difference from $\vec x$ best aligns with $\nabla E(\vec x)$; \ie, the update
becomes
\[
\vec x^\prime \gets \arg\max_{\vec z \in \mathcal{N}(\vec x)}{\tfrac{(\vec z-\vec x)}{\|\vec z-\vec x\|}^\top \nabla E(\vec x)} \enspace.
\]
Thus, the solution to the above alignment is simply to modify a
feature that satisfies $\arg\max_{i}| \nabla E(\vec x)_i|$ for which the
corresponding change leads to a feasible state.
%BB This section has to be revised: it is not true that it is enough to align with the best possible change. If the allowed step in the corresponding direction is too large, we may end up in increasing the objective function (basically, we jump off from the local minimum...). Therefore, a better strategy may be to consider one-by-one the best changes, and only accept the first that decreases the objective function.
%%%% BB I added this paragraph. Blaine, you may shorten / revise it a bit.
Note however that, sometimes, such a step may be relatively quite large, and may lead the attack out of a local minimum potentially increasing the objective function.
Therefore, one should consider the best alignment that effectively reduces the objective function by disregarding features that lead to states where the objective function is higher.

\subsection{Experiments}
\label{sect:exp-evasion-test-time}

In this section, we first report some experimental results on the
MNIST handwritten digit classification task
\cite{globerson-ICML06,LeCun95}, that visually demonstrate how the
proposed algorithm modifies digits to mislead classification. This
dataset is particularly useful because the visual nature of the
handwritten digit data provides a \emph{semantic meaning} for attacks.
We then show the effectiveness of the proposed attack on a more
realistic and practical scenario: the detection of malware in PDF
files.

\subsubsection{Handwritten Digits}
\label{sect:toy-example}

We first focus on a two-class sub-problem of discriminating between
two distinct digits from the MNIST dataset~\cite{LeCun95}.
Each digit example is represented as a
gray-scale image of $28 \times 28$ pixels
arranged in raster-scan-order to give feature vectors of $d = 28 \times 28 = 784$ values.  We normalized each feature (pixel)
$x_f \in [0,1]^{d}$ by dividing its value by $255$, and we
constrained the attack samples to this range.  Accordingly, we
optimized Equation~\eqref{eq:obj-function}
subject to $0 \leq x_f \leq 1$ for all $f$.
%  Each digit in the MNIST
%dataset is represented as a gray-scale image of $28
%\times 28$ pixels. The
%raster-scan-ordered pixel values of the image are the feature vector
%of this image.  The overall number of features is thus $d = 28 \times
%28 = 784$.  We normalized each feature (pixel value) $\vec x \in [0,1]^{d}$
%by dividing its value by $255$, and constrained the attack samples to
%this range.  Accordingly, we minimized the objective function in
%Equation~\eqref{eq:obj-function} subject to $0 \leq \vec x \leq 1$.

For our attacker, we assume the perfect knowledge (PK) attack scenario.
We used the \emph{Manhattan} distance ($\ell_{1}$-norm) as the
distance function, $d$, both for the kernel density
estimator (\ie, a Laplacian kernel) and for the
constraint $d(\vec x,\vec x^{0}) \leq d_{\rm max}$ of Equation~\eqref{eq:constraint},
which bounds the total difference between the gray level
values of the original image $\vec x^{0}$ and the attack image $\vec x$. We used an upper bound of $d_{\rm max} = \frac{5000}{255}$ to limit the total change in the gray-level values to $5000$. At
each iteration, we increased the $\ell_{1}$-norm value of $\vec x-\vec x^{0}$ by
$\frac{10}{255}$, which is equivalent to increasing the difference in the gray level values by $10$. This is effectively the gradient step size.

For the digit discrimination task, we applied an SVM with the linear kernel and $C=1$. We randomly chose $100$ training samples and applied the attacks to a correctly-classified positive sample.

\begin{figure}[htbp]
\begin{center}
\includegraphics[width=0.9\textwidth]{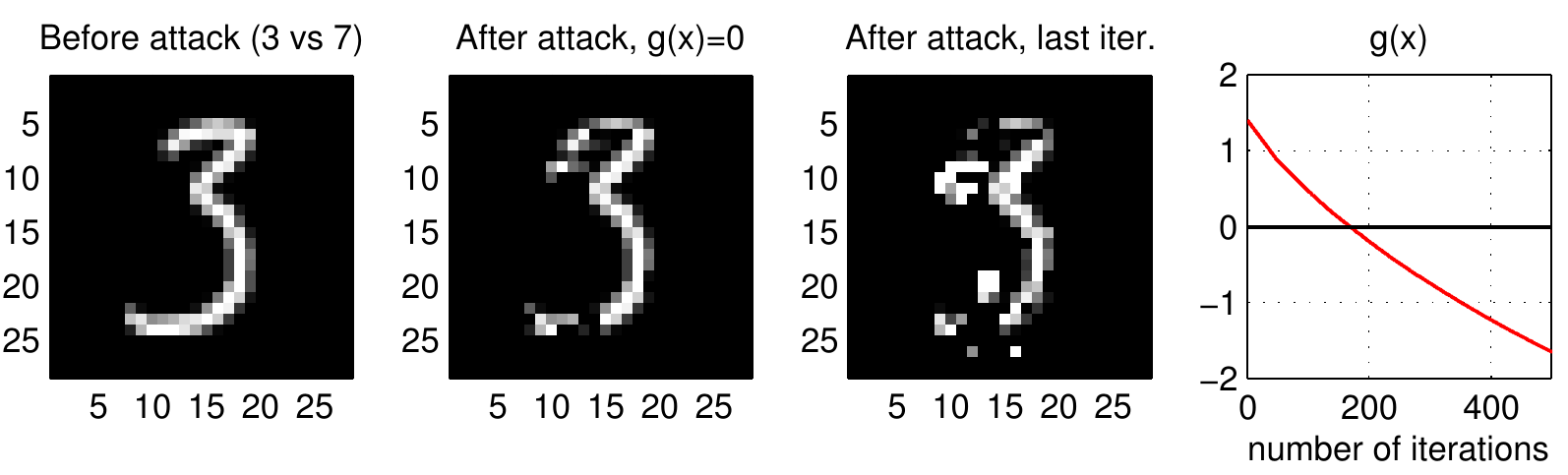}\\
\includegraphics[width=0.9\textwidth]{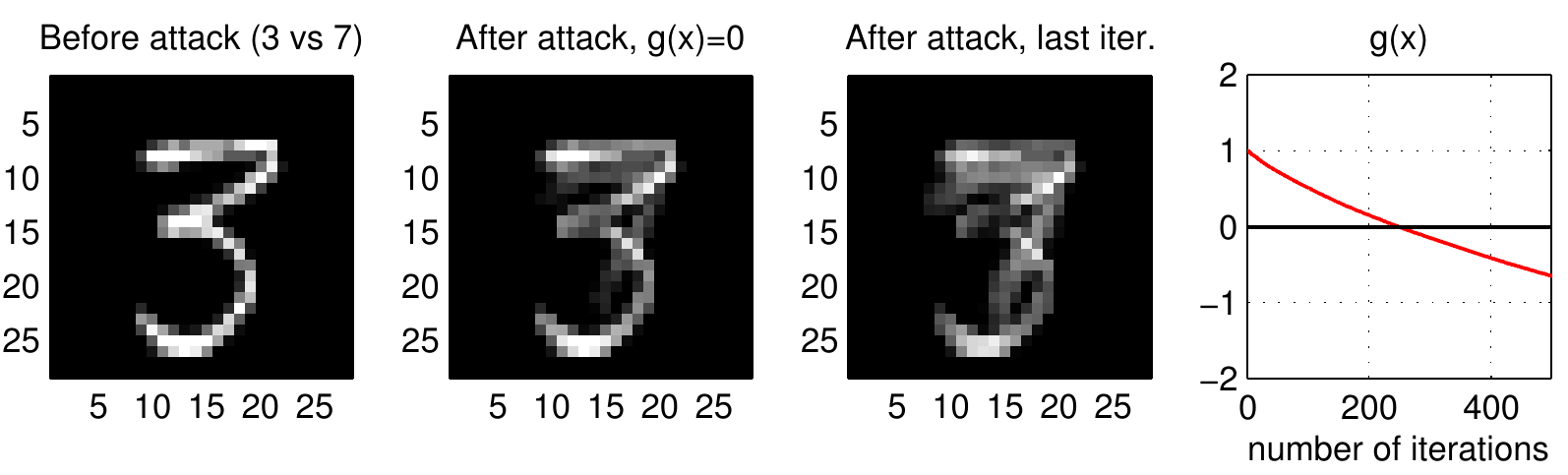}
\caption{Illustration of the gradient attack on the MNIST digit data, for $\lambda = 0$ (\textbf{top row}) and $\lambda=10$ (\textbf{bottom row}). Without a mimicry component ($\lambda = 0$) gradient descent quickly decreases $g$ but the resulting attack image does not resemble a ``7''. In contrast, the attack minimizes $g$ slower when mimicry is applied ($\lambda = 10$) but the final attack image closely resembles a mixture between ``3'' and ``7'', as the term ``mimicry'' suggests.}
%\caption{Illustration of the gradient attack on the MNIST digit data, for $\lambda = 0$ (top row) and $\lambda=10$ (bottom row).}
\label{fig:mnist}
\end{center}
\end{figure}

In Figure~\ref{fig:mnist} we illustrate gradient attacks in which a
``3'' is to be misclassified as a ``7''.
%The progress of gradient attacks, in which a ``3'' is to be
%misclassified as a ``7'', is illustrated in Figure~\ref{fig:mnist}.
The left image shows the initial attack point, the middle image shows
the first attack image misclassified as legitimate, and the right
image shows the attack point after 500 iterations.  When $\lambda=0$,
the attack images exhibit only a weak resemblance to the target class
``7'' but are, nevertheless, reliably misclassified. This is the same
effect we observed in the left plot of
Figure~\ref{fig:attack-strategy}: the classifier is evaded by making
the attack sample dissimilar to the malicious class.  Conversely, when
$\lambda=10$ the attack images strongly resemble the target class
because the mimicry term favors samples that are more similar to the
target examples. This is the same effect illustrated in the rightmost
plot of Figure~\ref{fig:attack-strategy}.

\subsubsection{Malware Detection in PDF Files}
\label{sect:malware-pdf}

We focus now on the problem of discriminating between legitimate and
malicious PDF files, a popular medium for disseminating
malware~\cite{IBM}. PDF files are excellent vectors for
malicious-code, due to their flexible \emph{logical structure}, which
can described by a hierarchy of interconnected objects. As a result,
an attack can be easily hidden in a PDF to circumvent file-type
filtering.  The PDF format further allows a wide variety of resources
to be embedded in the document including \texttt{JavaScript},
\texttt{Flash}, and even binary programs. The type of the embedded
object is specified by \emph{keywords}, and its content is in a
\emph{data stream}.
 Several recent works proposed machine-learning
techniques for detecting malicious PDFs use the file's logical
structure to accurately identify the malware~\cite{maiorca,Smutz,Srndic}.
In this case study, we use the feature representation
of Maiorca~\etal~\cite{maiorca} in which each feature corresponds to the tally of occurrences of a given keyword in the PDF file. Similar
feature representations were also exploited in~\cite{Smutz,Srndic}.

The PDF application imposes natural constraints on attacks.  Although
it is difficult to \emph{remove} an embedded object (and its
corresponding keywords) without corrupting the PDF's file
structure, it is rather easy to \emph{insert} new objects (and, thus,
keywords) through the addition of a new \emph{version} to the PDF
file~\cite{Refer}.  In our feature representation, this is equivalent
to allowing only feature increments,;\ie, requiring $\vec x^{0} \leq \vec x$ as
an additional constraint in the optimization problem given by
Equation~\eqref{eq:obj-function}.  Further, the total difference in keyword 
counts between two samples is their
\emph{Manhattan} distance, which we again use for the kernel
density estimator and the constraint in
Equation~\eqref{eq:constraint}. Accordingly, $d_{\rm max}$ is the
maximum number of additional keywords that an attacker 
can add to the original $\vec x^{0}$.

\medskip
\noindent \textbf{Experimental setup.} For experiments, we used a PDF corpus with
$500$ malicious samples from the \emph{Contagio}
dataset\footnote{\url{http://contagiodump.blogspot.it}} and $500$ benign
samples collected from the web. We randomly split the data into five
pairs of training and testing sets with $500$ samples each to average
the final results.
The features (keywords) were extracted from each
training set as described in \cite{maiorca};
on average, $100$ keywords were found in each run.
Further, we also bounded the maximum value of each feature (keyword count) to $100$, as this value was found to be close to the $95^{\rm th}$ percentile for each feature. This limited the influence of outlying samples having very high feature values.

We simulated the \emph{perfect} knowledge (PK) and the \emph{limited} knowledge (LK) scenarios described in Section~\ref{subsect:adv-knowledge-evasion-test-time}. In the LK case, we set the number of samples used to learn the surrogate classifier to $n_{q} =100$.
The reason is to demonstrate that even with a dataset as small as the $20\%$ of the original training set size, the adversary may be able to evade the targeted classifier with high reliability.
Further, we assumed that the adversary uses feedback from the \emph{targeted} classifier $f$; \ie, the labels $\hat y_{i} =  \classifierLBL[i] = f(\vec x_{i})$ for each surrogate sample $\vec x_{i}$. Similar results were also obtained using the true labels (without relabeling), since the targeted classifiers correctly classified almost all samples in the test set.
%We report results only for the case in which the
%adversary uses feedback from the \emph{targeted} classifier; \ie, the labels 
%$\classifierLBL$ for each surrogate sample.
%Similar results were also obtained with the true labels (without relabeling), since the
%targeted classifiers correctly classified almost all samples in the
%test set, and thus, relabeling gave nearly identical results.

As discussed in Section~\ref{sect:gradients-kde}, the value of $\lambda$ is chosen according to the scale of the discriminant function $g(\vec x)$, the bandwidth parameter $h$ of the kernel density estimator, and the number of samples $n$ labeled as legitimate in the surrogate training set.
For computational reasons, to estimate the value of the KDE at a given point $\vec x$ in the feature space, we only consider the $50$ nearest (legitimate) training samples to $\vec x$; therefore $n \leq 50$ in our case.
The bandwidth parameter was set to $h=10$, as this value provided a proper rescaling of the Manhattan distances observed in our dataset for the KDE. We thus set $\lambda = 500$ to be comparable with $O(nh)$.

For each targeted classifier and training/testing pair, we learned five different surrogate classifiers by randomly selecting $n_{q}$ samples from the test set, and averaged their results.
For SVMs, we sought a surrogate classifier that would correctly match the labels from the targeted classifier; thus, we used parameters $C=100$, and $\gamma=0.1$ (for the RBF kernel) to heavily penalize training errors. 

\begin{figure*}[ht]
\begin{center}
\includegraphics[width=0.48\textwidth]{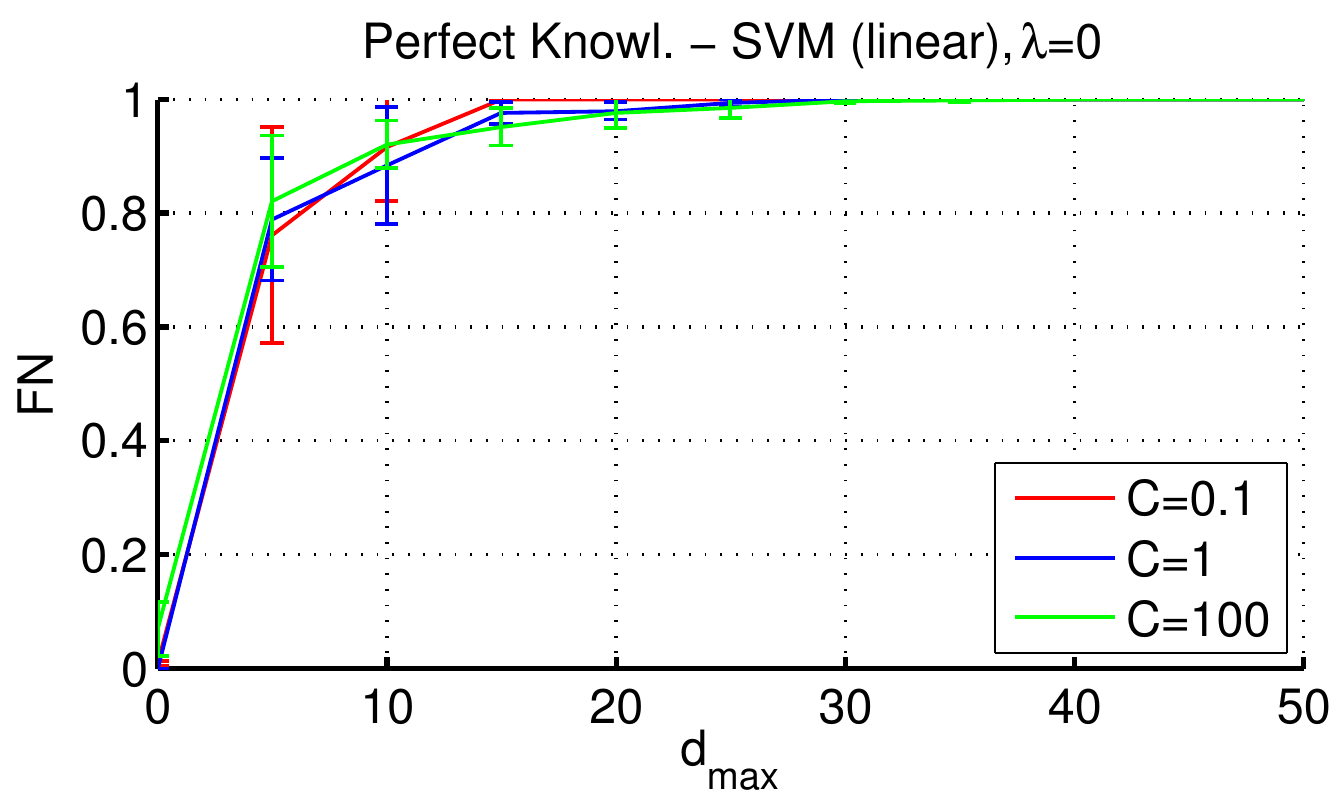}
\includegraphics[width=0.48\textwidth]{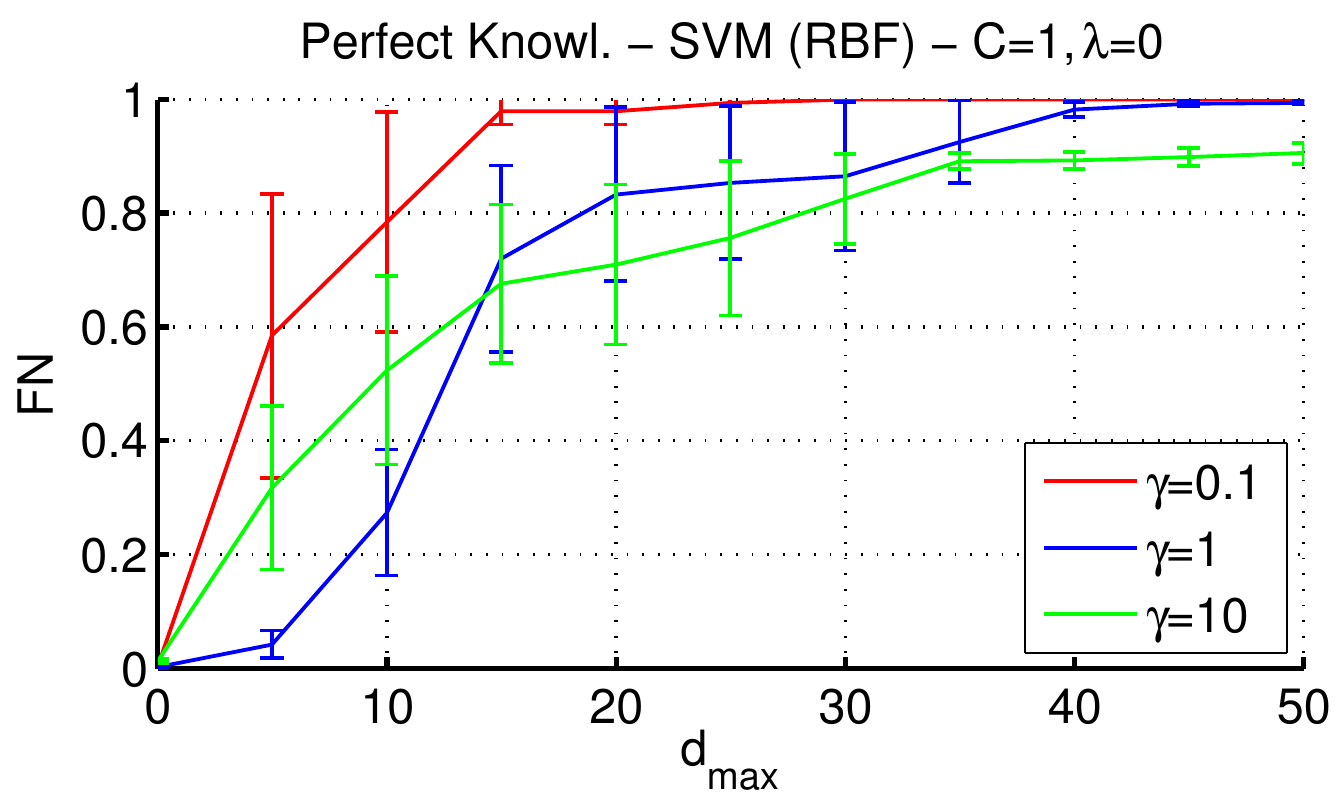}\\
\includegraphics[width=0.48\textwidth]{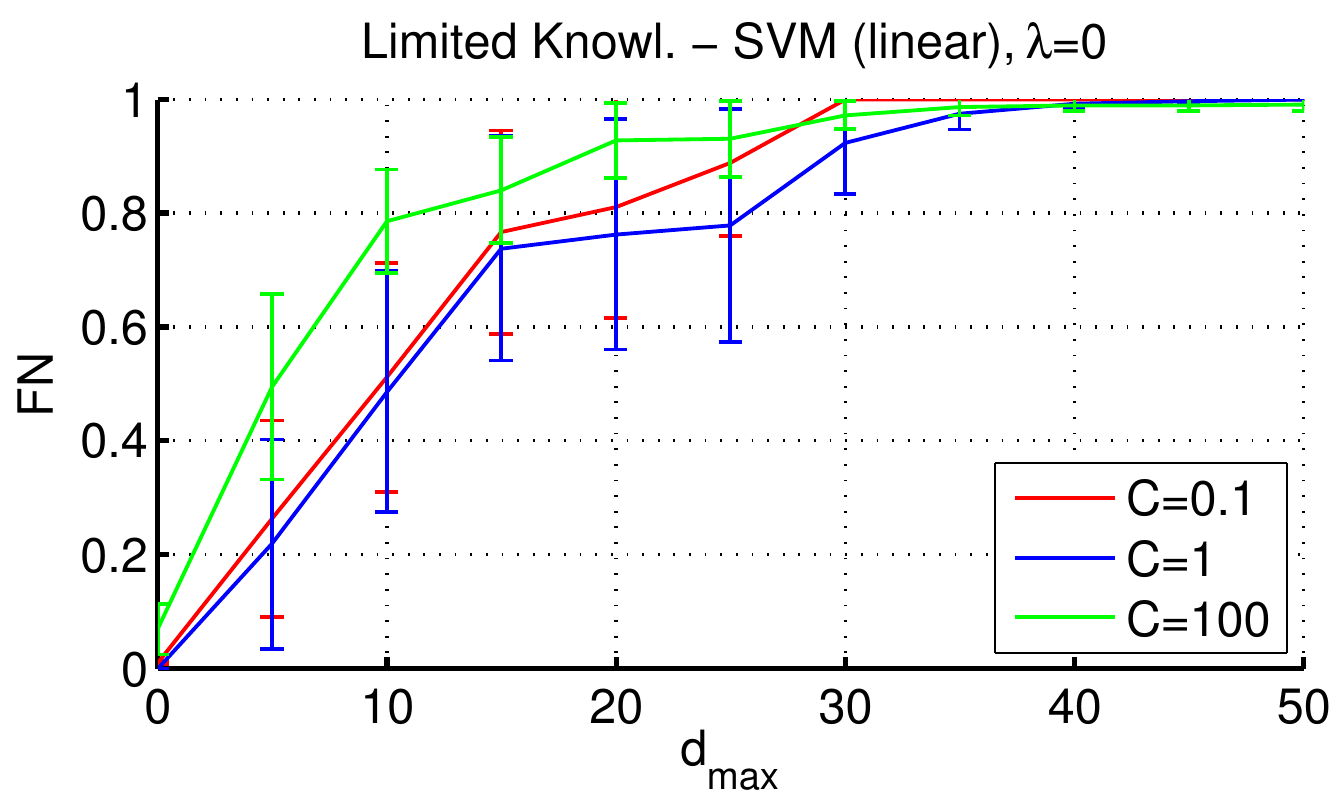}
\includegraphics[width=0.48\textwidth]{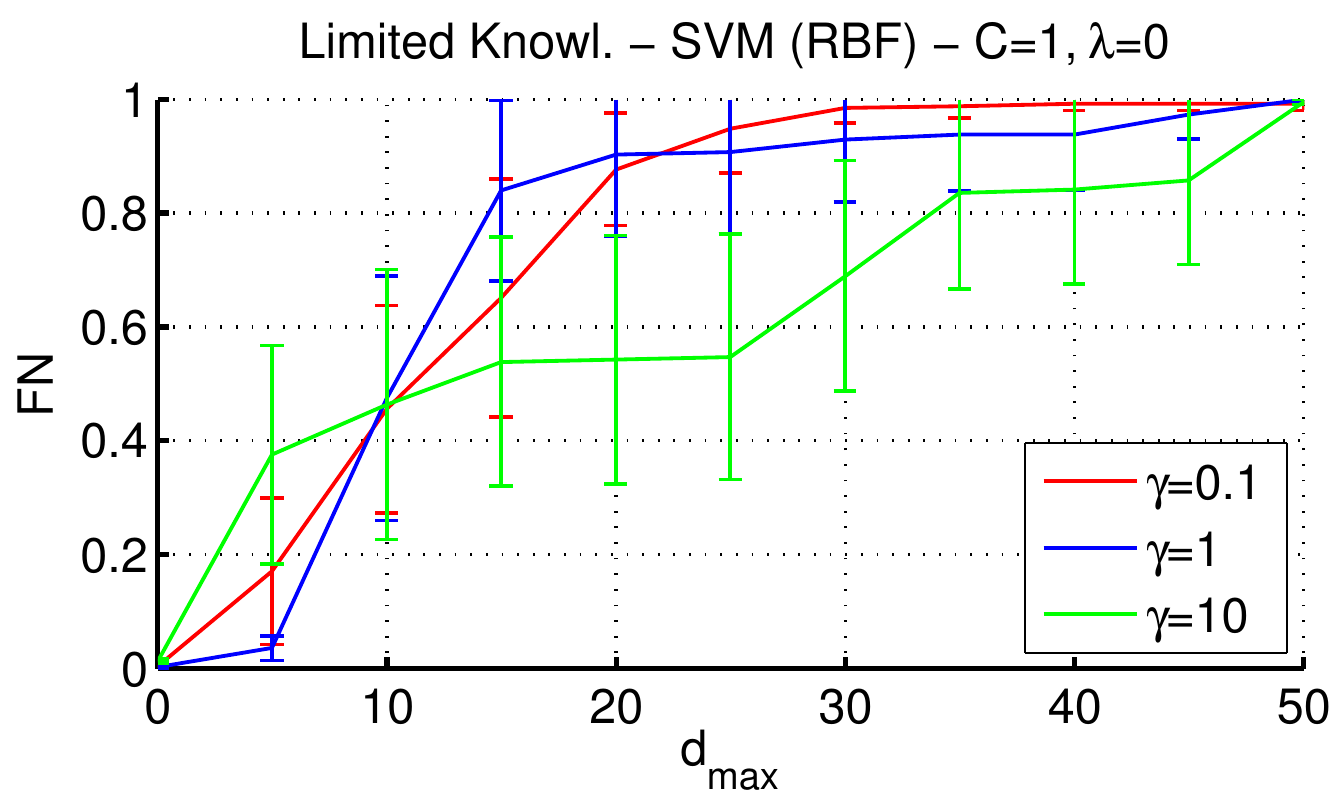} \\
\includegraphics[width=0.48\textwidth]{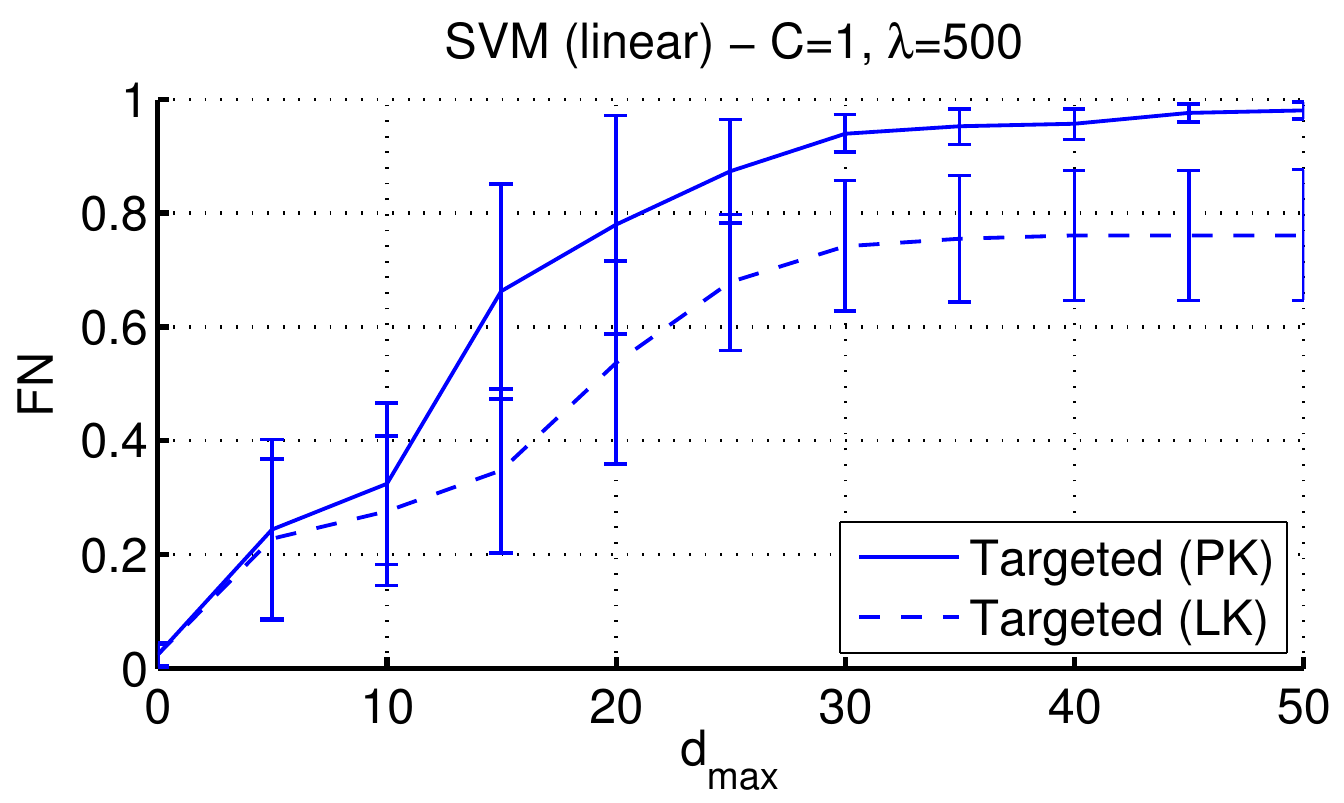}
\includegraphics[width=0.48\textwidth]{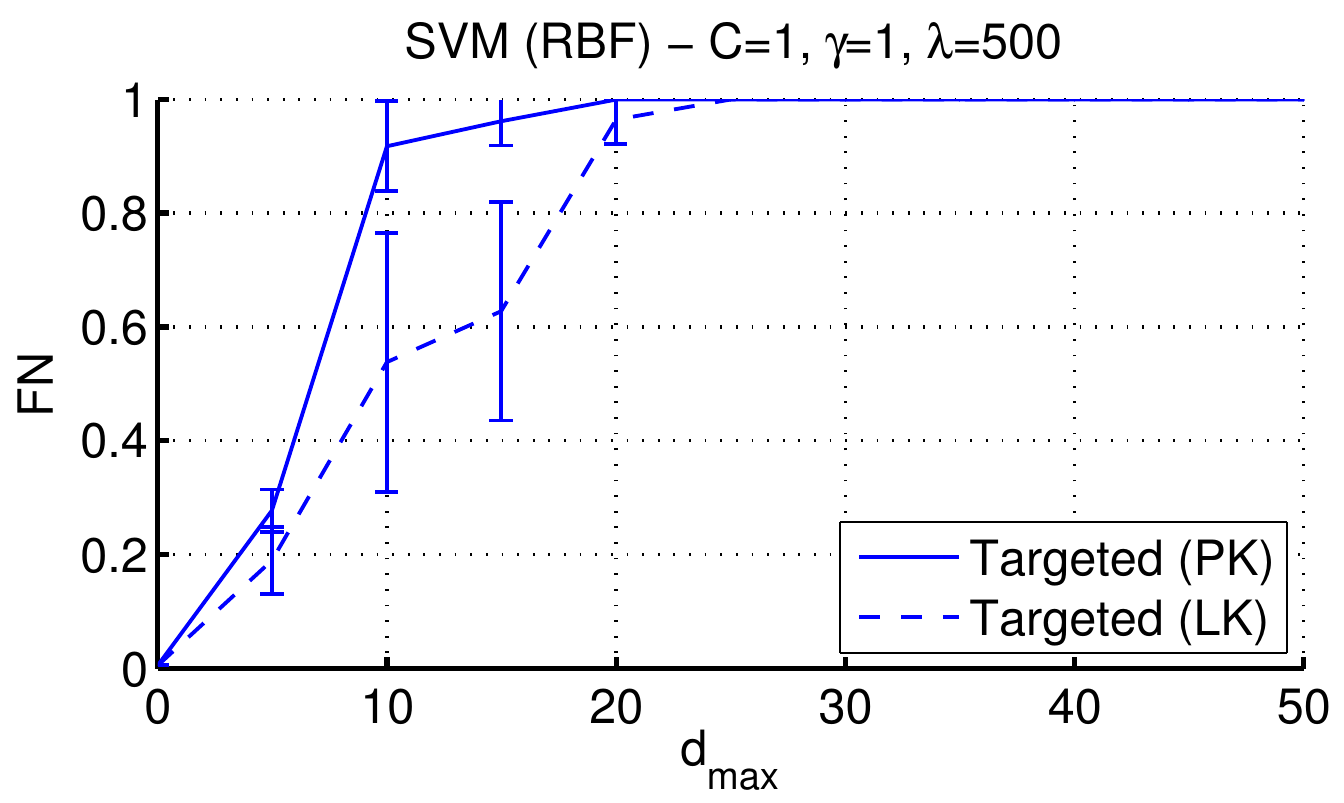}
\caption{Experimental results for SVMs with the linear and the RBF kernel
  (first and second column). We report the FN values (attained at
  FP=0.5\%) for increasing $d_{\rm max}$. For the sake of readability,
  we report the average FN value $\pm$ half standard deviation (shown
  with error bars).
  Results for perfect knowledge (PK) attacks with
  $\lambda=0$ (without mimicry) are shown in the first row, for
  different values of the considered classifier parameters, \ie, the
  regularization parameter $C$ for the linear SVM, and the kernel
  parameter $\gamma$ for the SVM with RBF kernel (with $C=1$).
  Results for limited
  knowledge (LK) attacks with $\lambda=0$ (without mimicry) are
  shown in the second row for the linear SVM (for varying $C$),
  and the SVM with RBF kernel (for varying $\gamma$, with $C=1$).
  Results for perfect (PK) and limited knowledge (LK) with
  $\lambda=500$ (with mimicry) are shown in the
  third row for the linear SVM (with $C=1$), and the SVM with RBF kernel (with $\gamma=1$ and $C=1$).}
\label{fig:results}
\end{center}
\end{figure*}

\medskip
\noindent \textbf{Experimental results.} We report our results in Figure~\ref{fig:results}, in terms of the
false negative (FN) rate attained by the targeted classifiers
as a function of the maximum allowable number of modifications, $d_{\rm max} \in [0, 50]$.
We compute the FN rate corresponding to a fixed false positive (FP) rate of FP$=0.5\%$. For $d_{\rm max}=0$, the FN rate
corresponds to a standard performance evaluation
using unmodified PDFs.
As expected, the FN rate increases with $d_{\rm max}$ as the PDF is increasingly modified, since the adversary has more flexibility in his attack. Accordingly, a more secure classifier will exhibit a more
graceful increase of the FN rate.

\smallskip
\noindent \textit{Results for $\lambda=0$.}
We first investigate the effect of the proposed attack in the PK case,
without considering the mimicry component (Figure~\ref{fig:results},
top row), for varying parameters of the considered classifiers.
The linear SVM (Figure~\ref{fig:results}, top-left plot) is almost
always evaded with as few as $5$ to $10$ modifications, independent of the
regularization parameter $C$.
It is worth noting that attacking a linear classifier
amounts to always incrementing the value of the same highest-weighted feature (corresponding to the \texttt{/Linearized} keyword in the majority of the cases) until it is bounded.
This continues with the next highest-weighted non-bounded feature until termination. This occurs simply because the gradient of $g(\vec x)$ does not depend on $\vec x$ for a linear classifier (see Section~\ref{subsubsect:gradient-svm}). %and this is a discrete feature space; thus, features are always modified in the same order for each attack sample. 
With the RBF kernel (Figure~\ref{fig:results}, top-right plot), SVMs
exhibit a similar behavior with $C=1$ and various values of its
$\gamma$ parameter,\footnote{We also conducted experiments using $C=0.1$ and
  $C=100$, but did not find significant differences compared to the
  presented results using $C=1$.} and the RBF SVM provides a higher degree of security compared
to linear SVMs (\emph{cf.} top-left plot and middle-left plot in
Figure~\ref{fig:results}).

In the LK case, without mimicry (Figure~\ref{fig:results}, middle
row), classifiers are evaded with a probability only \emph{slightly} lower than that found in the PK case, even
when only $n_{q}=100$ surrogate samples are used to learn the surrogate classifier. This aspect highlights the threat posed by a skilled adversary with incomplete knowledge: only a small set of samples may be required to successfully attack the target classifier using the proposed algorithm.

\smallskip
\noindent \textit{Results for $\lambda=500$.}
When mimicry is used (Figure~\ref{fig:results}, bottom row), the success of the evasion of linear SVMs (with $C=1$) decreases both in the PK
 (\eg, compare the blue curve in the top-left plot with the solid blue curve in the bottom-left plot)
and LK case
(\eg, compare the blue curve in the middle-left plot with the dashed blue curve in the bottom-left plot). The reason is that the computed direction tends to lead to a slower descent; \ie, a less direct path that often requires more modifications to evade the classifier.
In the non-linear case (Figure~\ref{fig:results}, bottom-right plot), instead, mimicking exhibits some beneficial aspects for the attacker, although the constraint on feature addition may make it difficult to properly mimic legitimate samples.
In particular, note how the targeted SVMs with RBF kernel (with $C=1$ and $\gamma=1$) in the PK case (\eg, compare the blue curve in the top-right plot with the solid blue curve in the bottom-right plot)
is evaded with a significantly higher probability than in the case when $\lambda=0$.
The reason is that a pure descent strategy on $g(\vec x)$ may find local minima (\ie, attack samples) that do not evade detection, while the mimicry component biases the descent towards regions of the feature space more densely populated by legitimate samples, where $g(\vec x)$ eventually attains lower values.
In the LK case (\eg, compare the blue curve in the middle-right plot with the dashed blue curve in the bottom-right plot), however, mimicking does not exhibit significant improvements.

\medskip
\noindent \textbf{Analysis.} 
Our attacks raise questions about the feasibility of detecting
malicious PDFs solely based on logical structure. We found that \texttt{/Linearized}, \texttt{/Open\-Ac\-tion}, \texttt{/Comment}, \texttt{/Root} and \texttt{/PageLayout} were among the most commonly manipulated keywords. They indeed are found mainly in
legitimate PDFs, but can be easily added to malicious PDFs by the
versioning mechanism. The attacker can simply insert comments inside
the malicious PDF file to augment its \texttt{/Comment} count.
Similarly, she can embed \emph{legitimate} OpenAction code to add
\texttt{/OpenAction} keywords or she can add new pages to insert
\texttt{/PageLayout} keywords.

In summary, our analysis shows that even detection systems that
accurately classify non-malicious data can be significantly degraded
with only a few malicious modifications. This aspect highlights the
importance of developing detection systems that are accurate, but also
\emph{designed to be robust} against adversarial manipulation of
attack instances.

\subsection{Discussion}
\label{subsect:discussion-evasion-test-time}

In this section, we proposed a simple algorithm that allows for evasion of SVMs with differentiable kernels, and, more generally, of any classifier with a differentiable discriminant function.
We investigated the attack effectiveness in the case of perfect knowledge of the attacked system. Further, we empirically showed that SVMs can still be evaded with high probability even if the adversary can only learn a classifier's copy on surrogate data (limited knowledge).
We believe that the proposed attack formulation can easily be extended to classifiers with non-differentiable discriminant functions as well, such as decision trees and $k$-nearest neighbors.

Our analysis also suggests some ideas for improving classifier
security.  In particular, when the classifier tightly \emph{encloses}
the legitimate samples, the adversary must increasingly mimic the
legitimate class to evade (see Figure~\ref{fig:attack-strategy}), and
this may not always be possible; \eg, malicious network packets or PDF
files still need to embed a valid exploit, and some features may be immutable.
%BB done some small modifications here as well (always be deliberately ...)
Accordingly, a guideline for designing secure classifiers is that
learning should encourage a tight enclosure of the legitimate class;
%the security of classifiers can be designed by considering
%regularization terms in their objective functions that promote
%enclosure of the legitimate class; 
\eg, by using a regularizer that penalizes classifying ``blind
spots''---regions with low $p(\vec x)$---as legitimate. Generative
classifiers can be modified, by explicitly modeling
the attack distribution, as in \cite{biggio11-smc}, and discriminative
classifiers can be modified similarly by adding generated attack
samples to the training set. However, these security improvements may incur higher FP rates.

In the above applications, the feature representations were
\emph{invertible}; \ie, there is a direct mapping from the feature
vectors $\vec x$ to a corresponding real-world sample (\eg, a spam email,
or PDF file).  However, some feature mappings can not be trivially
inverted; \eg, $n$-gram analysis \cite{fogla06}.  In these cases, one
may modify the real-world object corresponding to the initial attack
point at each step of the gradient descent to obtain a sample in the
feature space that as close as possible to the sample that would
be obtained at the next attack iteration. A similar technique has been
already exploited in to address the pre-image
problem of kernel methods~\cite{biggio12-icml}.
%, \ie, to find points in the input space that
%correspond to their images in the feature space.

Other interesting extensions include (i) considering more effective strategies such as those proposed by \cite{lowd05,nelson12-jmlr} to build a small but representative set of surrogate data to learn the surrogate classifier and (ii) improving the classifier estimate $\hat g(\vec x)$; \eg using an ensemble technique like bagging %or the random subspace method \cite{ho98}
 to average several classifiers~\cite{breiman96-bagging}.

\section{Poisoning Attacks against SVMs} %5-8 pages [Blaine]
\label{sect:poisoning}

In the previous section, we devised a simple algorithm that allows for evasion of classifiers at test time and showed experimentally how it can be exploited to evade detection by SVMs and kernel-based classification techniques. Here we present another kind of attack, based on our work in \cite{biggio12-icml}. Its goal is to force the attacked SVM to misclassify as many samples as possible at test time through \emph{poisoning} of the training data, that is, by injecting well-crafted attack samples into the training set. Note that, in this case, the test data is assumed not to be manipulated by the attacker.

Poisoning attacks are staged during classifier training, and they can thus target \emph{adaptive} or \emph{online} classifiers, as well as classifiers that are being re-trained on data collected during test time, especially if in an unsupervised or semi-supervised manner. Examples of these attacks, besides our work \cite{biggio12-icml}, can be found in \cite{biggio12-spr,biggio11-mcs,biggio12-icb,kloft07,kloft12b,nelson08,rubinstein09}. They include specific application examples in different areas, such as intrusion detection in computer networks \cite{biggio11-mcs,kloft07,kloft12b,rubinstein09}, spam filtering \cite{biggio11-mcs,nelson08}, and, most recently, even biometric authentication \cite{biggio12-icb,biggio12-spr}.

In this section, we follow the same structure of Section~\ref{sect:evasion}.
In Section~\ref{sect:adv-model-poisoning}, we define the adversary model according to our framework;
then, in Sections~\ref{sect:poisoning-attack-strategy} and~\ref{sect:poisoning-attack-algorithm} we respectively derive the optimal poisoning attack and the corresponding algorithm; and, finally, in Sections~\ref{sect:poisoning-experiments} and~\ref{sect:poisoning-discussions} we report our experimental findings and discuss the results. %the reader probably expects some experiments on an application example, as the malware PDF in the previous section. It would be thus better to point out here that we will only present experiments on handwritten digit recognition.

\subsection{Modeling the Adversary}
\label{sect:adv-model-poisoning}

Here, we apply our framework to evaluate security against poisoning
attacks. As with the evasion attacks in
Section~\ref{sect:adv-model-evasion-test-time}, we model the attack
scenario and derive the corresponding optimal attack strategy for
poisoning.

\textbf{Notation}. In the following, we assume that an SVM has been trained on a dataset $\mathcal
D_{\text{tr}} = \{ \vec x_{i}, y_{i} \}_{i=1}^{n}$ with $\vec x_i \in
\R^d$ and $y_{i} \in \{-1,+1\}$.  The matrix of kernel values between two sets of points is denoted with $\vec K$, while $\vec Q = \vec K \circ yy^{\top}$ denotes
its label-annotated version, and $\alpha$ denotes the SVM's dual
variables corresponding to each training point.  Depending on the value
of $\alpha_i$, the training points are referred to as margin support
vectors ($0 < \alpha_i < C$, set $\mathcal{S}$), error support vectors
($\alpha_i = C$, set $\mathcal{E}$) or reserve vectors ($\alpha_i =
0$, set $\mathcal{R}$). In the sequel, the lower-case letters $s,e,r$ are used to index the
corresponding parts of vectors or matrices; \eg, $\vec Q_{ss}$ denotes the sub-matrix of $\vec Q$ corresponding to 
the margin support vectors.

\subsubsection{Adversary's Goal}

For a poisoning attack, the attacker's goal is to find a set of points whose addition to $\mathcal D_{\text{tr}}$ maximally decreases the SVM's classification accuracy.
For simplicity, we start considering the addition of a single attack point $(\vec x^{*},y^{*})$. 
The choice of its label $y^{*}$ is arbitrary but fixed. We refer to the class of
this chosen label as \emph{attacking} class and the other as the
\emph{attacked} class.

\subsubsection{Adversary's Knowledge}

According to Section~\ref{sect:adversary-knowledge}, we assume that the adversary knows the training samples \knowref{know-1}, the feature representation \knowref{know-2}, that an SVM learning algorithm is used \knowref{know-3}
and the learned SVM's parameters \knowref{know-4}, since they can be inferred by the adversary by solving the SVM learning problem on the \emph{known} training set. Finally, we assume that no feedback is exploited by the adversary \knowref{know-5}.

These assumptions amount to considering a worst-case analysis that
allows us to compute the maximum error rate that the adversary can
inflict through poisoning.  This is indeed useful to check whether and
under what circumstances poisoning may be a relevant threat
for SVMs.

Although having perfect knowledge of the training data is very difficult in practice for an adversary, collecting a surrogate dataset sampled from the same distribution may not be that complicated; for instance, in network intrusion detection an attacker may easily sniff network packets to build a surrogate learning model, which can then be poisoned under the perfect knowledge setting. The analysis of this limited knowledge poisoning scenario is however left to future work.

\subsubsection{Adversary's Capability}
According to Section~\ref{sect:adversary-capability}, we assume that the attacker can manipulate only training data \capref{cap-1}, can manipulate the class prior and the class-conditional distribution of the attack point's class $y^{*}$ by essentially adding a number of attack points of that class into the training data, one at a time \capref[cap-2]{cap-3}, and can alter the feature values of the attack sample within some lower and upper bounds \capref{cap-4}. In particular, we will constrain the attack point to lie within a box, that is $\vec x_{\rm lb} \leq \vec x \leq \vec x_{\rm ub}$.

\subsubsection{Attack Strategy}
\label{sect:poisoning-attack-strategy}

Under the above assumptions, the optimal attack strategy amounts to solving the following optimization problem:
\begin{eqnarray} 
\label{eq:L_def}
\vec x^{*} =  & {\rm argmax}_{\vec x} \enspace P(\vec x) = \sum_{k=1}^{m} (1 - y_k f_{\vec x}(\vec x_k))_+ & =\sum_{k=1}^{m} (-g_k)_+ \enspace \\
\label{eq:poisoning-constraint}
 & {\rm s.t.} \enspace \vec x_{\rm lb} \leq \vec x \leq \vec x_{\rm ub}  \enspace, &
\end{eqnarray}
where the hinge loss has to be maximized on a separate validation set $\mathcal D_{\text{val}} = \{
\vec x_{k}, y_{k} \}_{k=1}^{m}$ to avoid considering a further regularization term in the objective function.
The reason is that the attacker aims to maximize the SVM \emph{generalization error} and not only its empirical estimate on the training data.

\subsection{Poisoning Attack Algorithm}
\label{sect:poisoning-attack-algorithm}

In this section, we assume the role of the attacker and develop a
method for optimizing $\vec x^{*}$ according to Equation~\eqref{eq:L_def}.
Since the objective function is non-linear, we use a gradient-ascent algorithm, where the attack vector
is initialized by cloning an arbitrary point from the attacked class
and flipping its label.  This initialized attack point (at iteration
0) is denoted by $\vec x^{0}$.  In principle, $\vec x^{0}$ can be any point
\emph{sufficiently deep} within the attacking class's margin. However,
if this point is too close to the boundary of the attacking class, the
iteratively adjusted attack point may become a reserve point, which
halts further progress.

\begin{algorithm}[tb]
  \caption{Poisoning attack against an SVM}
  \label{alg:poison}
  \textbf{Input:} $\mathcal D_{\rm tr}$, the training data; $\mathcal D_{\rm val}$, the validation data; $y^{*}$, the class label of the attack point; $\vec x^{0}$, the initial attack point; $t$, the step size.\\
  \textbf{Output:} $\vec x^{*}$, the final attack point.
  
  \begin{algorithmic}[1]
        \STATE{ $\{ \alpha_{i}, b \} \leftarrow $ learn an SVM on $\mathcal D_{\rm tr}$.}
    \STATE{$p \gets 0$.}
    \REPEAT
            \STATE{Re-compute the SVM solution on $\mathcal D_{\rm tr} \cup \{\vec x^{p}, y^{*} \}$ using the incremental SVM \cite{cauwenberghs00}. This step requires $\{ \alpha_{i}, b \}$. If computational complexity is manageable, a full SVM can be learned at each step, instead.}
     \STATE{Compute $\nabla{P}$ on $\mathcal{D}_{\rm val}$ according to
      Equation~\eqref{eq:L_grad}.}
      \STATE{Normalize $\nabla{P}$ to have unit norm.} 
      \STATE{$p \gets p + 1$ and $\vec x^{p} \gets  \vec x^{p-1} + t \nabla{P}$}
      \IF{$\vec x_{\rm lb} > \vec x^{p}$ or $\vec x^{p} > \vec x_{\rm ub}$ }
     \STATE{Project $\vec x^{p}$ onto the boundary of the feasible region (enforce application-specific constraints, if any).}
     \ENDIF
      \UNTIL{$P\left(\vec x^{p}\right) - P\left( \vec x^{p-1}\right) < \epsilon$}
      \STATE{\textbf{return:} $\vec x^{*} = \vec x^{p}$}
  \end{algorithmic}
\end{algorithm}

The computation of the gradient of the validation error crucially
depends on the assumption that the structure of the sets
$\mathcal{S}$, $\mathcal{E}$ and $\mathcal{R}$ does not change during
the update. In general, it is difficult to determine the largest step
$t$ along the gradient direction $\nabla P$, which preserves this structure.
Hence, the step $t$ is fixed to a small constant value in our algorithm. After
each update of the attack point $\vec x^{p}$, the optimal solution can be 
efficiently recomputed from the solution on $\mathcal{D}_{\rm tr}$,
using the incremental SVM machinery \cite{cauwenberghs00}.
The algorithm terminates when the change in the validation error is smaller than a predefined threshold.

\subsubsection{Gradient Computation}

We now discuss how to compute the gradient $\nabla P$ of our objective
function.  For notational convenience, we now refer to
the attack point as $\vec x_{c}$ instead of $\vec x$.

First, we explicitly account for all terms in the margin conditions
$g_k$ that are affected by the attack point $\vec x_{c}$:
\begin{align}
    g_k &= \sum_{j} \vec Q_{kj} \alpha_j + y_k b - 1 \label{eq:gk_def}\\
    &= \sum_{j \neq c} \vec Q_{kj} \alpha_j (\vec x_c) +
      \vec Q_{kc}(\vec x_c) \alpha_c (\vec x_c) + y_k b(\vec x_c) - 1 \enspace. \nonumber
    \end{align}
As already mentioned, $P(\vec x_{c})$ is a non-convex objective function, and we thus exploit a gradient ascent technique to iteratively optimize it.
We denote the initial location of the attack point as $\vec x_c^{0}$.
% (where we added the subscript $c$ for notational convenience).
Our goal is to update the attack point as $\vec x_c^{p} =  \vec x_c^{(p-1)} + t \nabla P$ where $p$ is the current iteration, $\nabla P$ is a unit vector representing the attack direction (\ie, the normalized objective gradient), and $t$ is the step size. To maximize our objective, the attack direction $\nabla P$ is computed at each iteration.  

Although the hinge loss is not everywhere differentiable,
this can be overcome by only considering point
indices $k$ with non-zero contributions to
$P$; \ie, those for which $-g_k > 0$. Contributions of such points to
$\nabla P$ can be computed by differentiating
Equation~\eqref{eq:gk_def} with respect to $\vec x_{c}$ using the product rule:

\begin{equation}
  \label{eq:gk_diff}
  \diff{g_k}{\vec x_{c}} = \vec Q_{ks} \diff{\alpha}{\vec x_{c}} +
  \diff{\vec Q_{kc}}{\vec x_{c}} \alpha_c + y_k \diff{b}{\vec x_{c}},
\end{equation}
where, by denoting the ${l}^{\rm th}$ feature of $\vec x_{c}$ as $x_{cl}$, we use the notation
\begin{equation*}
  \diff{\alpha}{\vec x_{c}} = 
  \begin{bmatrix}
    \diff{\alpha_1}{x_{c1}} & \cdots & \diff{\alpha_1}{x_{cd}} \\
    \vdots & \ddots & \vdots\\
    \diff{\alpha_s}{x_{c1}} & \cdots & \diff{\alpha_s}{x_{cd}}
  \end{bmatrix}, \;\text{simil.}\; \diff{\vec Q_{kc}}{\vec x_{c}},\, \diff{b}{\vec x_{c}} \enspace .
\end{equation*}

The expressions for the gradient can be further refined using the fact
that the gradient step must preserve the optimal SVM solution. This
can expressed as an adiabatic update condition using the technique
introduced in \cite{cauwenberghs00}. In particular, for the ${i}^{\rm
  th}$ \emph{training} point, the KKT conditions of the optimal
SVM solution are:

\begin{align}
g_{i} &=   \sum_{j \in \mathcal{D}_{\text{tr}}} \vec Q_{ij} \alpha_{j}+y_{i}b-1
\begin{cases}
> 0; \;  i \in \mathcal R\\
= 0; \;  i \in \mathcal S \\
< 0; \;  i \in \mathcal E
\end{cases} \label{eq:kt-svm-1}\\
h &=  \tsum_{j \in \mathcal{D}_{\text{tr}}} \left(y_{j}\alpha_{j}\right)=0 \label{eq:kt-svm-2} \enspace .
\end{align}
The form of these conditions implies that an infinitesimal change in the attack
point $x_c$ causes a smooth change in the optimal solution of the
SVM, under the restriction that the composition of the sets
$\mathcal{S}$, $\mathcal{E}$ and $\mathcal{R}$ remains intact. This
equilibrium allows us to predict the \emph{response} of the SVM solution
to the variation of $x_c$, as shown below.

By differentiation of the $x_c$-dependent terms in
Equations~\eqref{eq:kt-svm-1}--\eqref{eq:kt-svm-2} with respect to each
feature $x_{cl}$ ($1 \leq l \leq d$), we obtain, for any $i \in \mathcal S$,
\begin{equation*}
  %\label{eq:kt_diff}
  \begin{aligned}
  \diff{g}{x_{cl}} &= \vec Q_{ss} \diff{\alpha}{x_{cl}} + \diff{\vec Q_{sc}}{x_{cl}} \alpha_c +
    y_s \diff{b}{x_{cl}} = 0\\
    \diff{h}{x_{cl}} &= y_s^\T \diff{\alpha}{x_{cl}} = 0 \enspace .
  \end{aligned}
\end{equation*}
%which can be rewritten as
Solving these equations and computing an inverse matrix via the Sherman-Morrison-Woodbury formula~\cite{Lue96} yields the following gradients:
%\begin{equation}
%  \label{eq:kt_diff_mat}
%  \begin{aligned}
%  \begin{bmatrix}
%    \diff{b}{x_{cl}} \\ \diff{\alpha}{x_{cl}}
%  \end{bmatrix} &=  -
%  {\begin{bmatrix}
%    0 & y_S^\T \\
%    y_s & \vec Q_{ss}
%  \end{bmatrix}}^{-1}
%  \begin{bmatrix}
%    0 \\ \diff{\vec Q_{sc}}{x_{cl}}
%  \end{bmatrix}
%  \alpha_c 
%  \enspace .
%  \end{aligned}
%\end{equation}
%The first matrix can be inverted using the
%Sherman-Morrison-Woodbury formula~\cite{Lue96}:
%\begin{equation}
%  \label{eq:smw}
%  \begin{aligned}
%    \begin{bmatrix}
%      0 & y_s^{\T} \\
%      y_s & \vec Q_{ss}
%    \end{bmatrix}^{-1}
%    = \frac{1}{\zeta}
%    \begin{bmatrix}
%      -1 & \upsilon^\T \\
%      \upsilon & \zeta \vec Q_{ss}^{-1} - \upsilon \upsilon^{\T}
%    \end{bmatrix}
%  \end{aligned}
%\end{equation}
%where $\upsilon = \vec Q_{ss}^{-1} y_s$ and $\zeta = y_s^\T \vec Q_{ss}^{-1}y_s$.
%Substituting Equation~\eqref{eq:smw} into Equation~\eqref{eq:kt_diff_mat} and observing
%that all components of the inverted matrix are independent of $x_c$, we obtain:
\begin{equation*}
  %\label{eq:adiabat}
  \begin{aligned}
    \diff{\alpha}{\vec x_{c}} &= - \frac{1}{\zeta} \alpha_c (\zeta \vec Q_{ss}^{-1} - \upsilon
    \upsilon^\T) \cdot \diff{\vec Q_{sc}}{\vec x_{c}} \\
    \diff{b}{\vec x_{c}} &= - \frac{1}{\zeta} \alpha_c  \upsilon^\T \cdot \diff{\vec Q_{sc}}{\vec x_{c}} \enspace,
  \end{aligned}
\end{equation*}
where $\upsilon = \vec Q_{ss}^{-1} y_s$ and $\zeta = y_s^\T \vec Q_{ss}^{-1}
y_s$.
%Substituting Equation~\eqref{eq:adiabat}) into Equation~\eqref{eq:gk_diff} and further
%into Equation~\eqref{eq:L_def}, 
We thus obtain the following gradient of the objective used for optimizing our attack, which only depends on $\vec x_c$ through gradients of the kernel matrix, $\diff{\vec Q_{kc}}{\vec x_{c}}$:
\begin{equation}
  \label{eq:L_grad}
  \nabla P = \sum_{k=1}^m  \left \{   M_k
  \diff{\vec Q_{sc}}{\vec x_{c}} +  \diff{\vec Q_{kc}}{\vec x_{c}} \right \} \alpha_{c} \enspace, 
\end{equation}
where $M_k = -\frac{1}{\zeta} ( \vec Q_{ks} (\zeta \vec Q_{ss}^{-1} - \upsilon\upsilon^{T})  + y_{k} \upsilon^{T})$.

\subsubsection{Kernelization}

From Equation \eqref{eq:L_grad}, we see that the gradient of the objective function at iteration $p$ may depend on the attack point $\vec x_{c}^{p}=\vec x_{c}^{p-1}+t \nabla P$ only through the gradients of the matrix $\vec Q$. In particular, this depends on the chosen kernel. We report below the expressions of these gradients for three common kernels (see Section~\ref{subsubsect:gradient-svm}):

\begin{compactitem}
\item Linear kernel:
  $
  \diff{K_{ic}}{\vec x_{c}} = \diff{(\vec x_i \cdot \vec x_c)}{\vec x_{c}} = \vec x_i
  $
\item Polynomial kernel:
  $
  \diff{K_{ic}}{\vec x_{c}} = \diff{(\vec x_i \cdot \vec x_c+ R)^d}{\vec x_{c}} = d(\vec x_i \cdot \vec x_{c} + R)^{d-1} \vec x_i 
  $
\item RBF kernel:
  $
  \diff{K_{ic}}{\vec x_{c}} = \diff{e^{-\gamma ||\vec x_i - \vec x_c||^2}}{\vec x_{c}} = 2\gamma K(\vec x_i,\vec x_c)  (\vec x_i -\vec  x_c)
  $
\end{compactitem}

The dependence of these gradients on the current
attack point, $\vec x_{c}$, can be avoided by using the previous attack
point, provided that $t$ is sufficiently small. This approximation
enables a straightforward extension of our method to arbitrary
differentiable kernels.

\subsection{Experiments}
\label{sect:poisoning-experiments}

The experimental evaluation presented in the following sections
demonstrates the behavior of our proposed method on an artificial
two-dimensional dataset and evaluates its effectiveness on the
classical MNIST handwritten digit recognition dataset~\cite{globerson-ICML06,LeCun95}.

\subsubsection{Two-dimensional Toy Example}

Here we consider a two-dimensional example in which each
class follows a Gaussian with mean and covariance
matrices given by $\mu_{-} = [-1.5, 0]$, $\mu_{+} = [1.5, 0]$, $\Sigma_{-}
= \Sigma_{+} = 0.6 I$. The points from the \emph{negative} distribution
have label $-1$ (shown as red in subsequent figures)
and otherwise $+1$ (shown as blue).  The training and the validation sets,
$\mathcal D_{\rm tr}$ and $\mathcal D_{\rm val}$, consist of $25$
and $500$ points per class, respectively.

\begin{figure*}[tb]
\begin{center}
\includegraphics[width=0.45\textwidth]{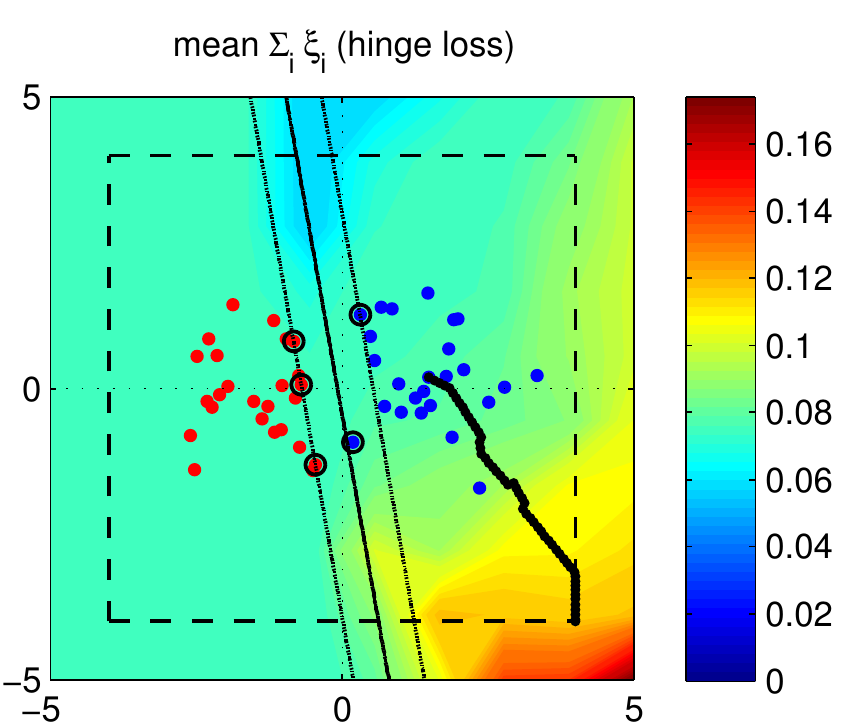}
\includegraphics[width=0.45\textwidth]{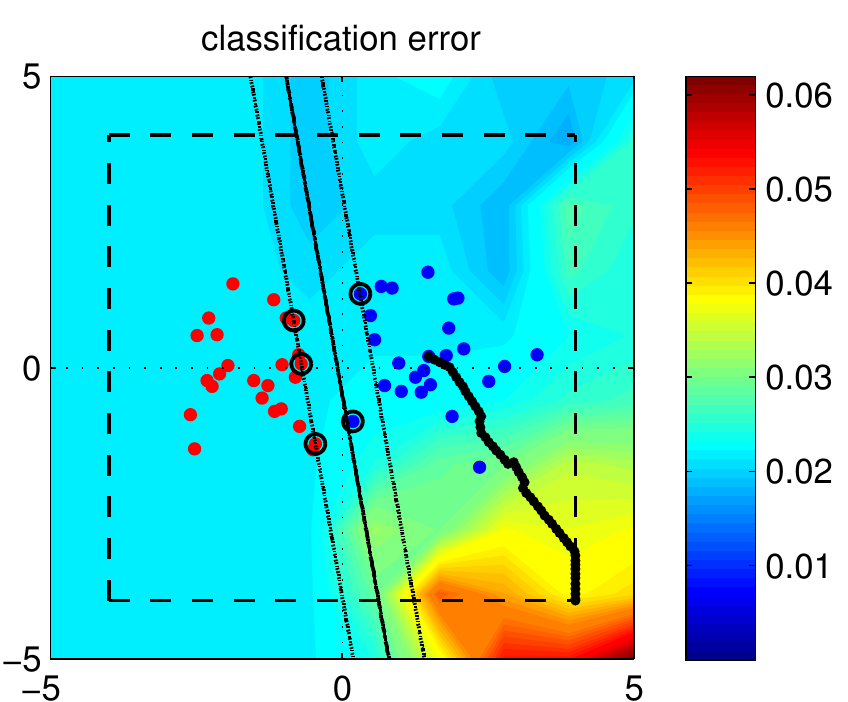}\\
\includegraphics[width=0.45\textwidth]{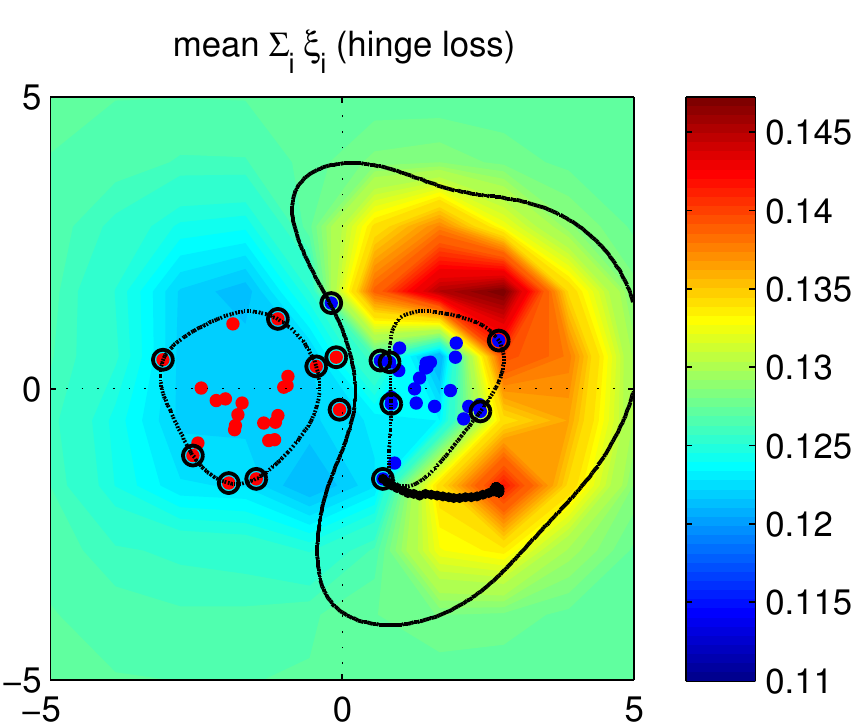}
\includegraphics[width=0.45\textwidth]{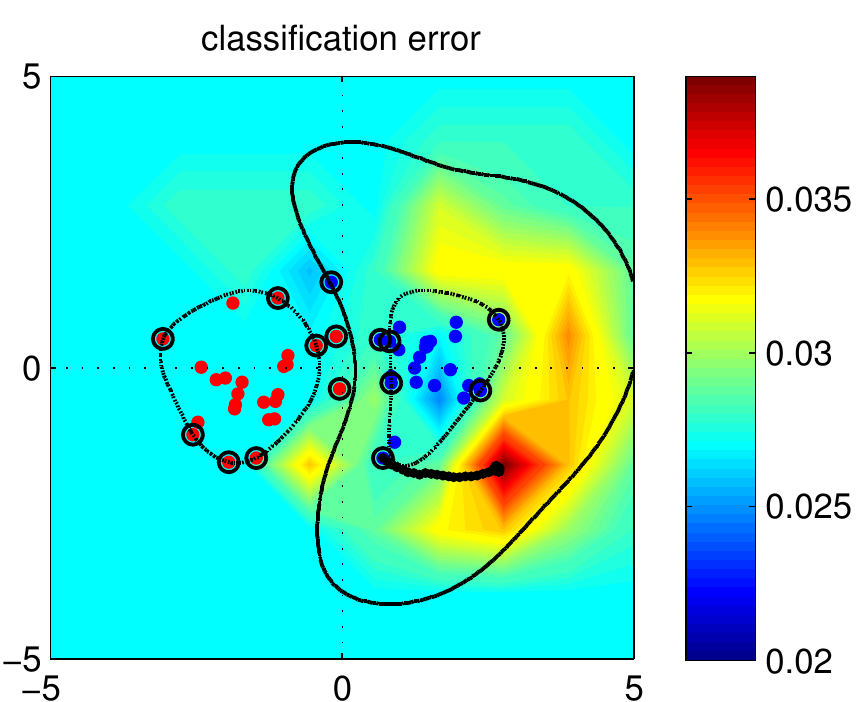}
\caption{Behavior of the gradient-based attack strategy on the
  Gaussian datasets, for the linear (top row) and the RBF kernel
  (bottom row) with $\gamma=0.5$. The regularization parameter $C$ was
  set to $1$ in both cases.  The solid black line represents the gradual
  shift of the attack point $\vec x_{c}^{p}$ toward a local maximum.  The hinge
  loss and the classification error are shown in colors, to appreciate
  that the hinge loss provides a good approximation of the
  classification error. The value of such functions for each point $\vec x
  \in [-5,5]^{2}$ is computed by learning an SVM on $\mathcal{D}_{\rm tr} \cup
  \{\vec x,y=-1\}$ and evaluating its performance on $\mathcal{D}_{\rm
    val}$. The SVM solution on the clean data $\mathcal{D}_{\rm tr}$,
  and the training data itself, are reported for completeness,
  highlighting the support vectors (with black circles), the 
  decision hyperplane and the  margin bounds (with black lines).}
\label{fig:exp-artificial}
\end{center}
\end{figure*}

In the experiment presented below, the red class is the
\emph{attacking} class. That is, a random point of the blue class is
selected and its label is flipped to serve as the starting point for
our method.  Our gradient ascent method is then used to refine this
attack until its termination condition is satisfied. The attack's
trajectory is traced as the black line in
Figure~\ref{fig:exp-artificial} for both the linear kernel (upper two
plots) and the RBF kernel (lower two plots). The background of each
plot depicts an error surface: hinge loss computed on a validation set
(leftmost plots) and the classification error (rightmost plots). For
the linear kernel, the range of attack points is limited to the box $\vec x
\in [-4,4]^{2}$ shown as a dashed line. This implements the constraint
of Equation~\eqref{eq:poisoning-constraint}.

For both kernels, these plots show that our gradient ascent
algorithm finds a reasonably good local maximum of the
non-convex error surface. For the linear kernel, it terminates at the
corner of the bounded region, since the error surface is
unbounded. For the RBF kernel, it also finds a good local maximum of the
hinge loss which, incidentally, is the maximum classification error within
this area of interest. 

\subsubsection{Handwritten Digits}

We now quantitatively validate the effectiveness of the proposed
attack strategy on the MNIST handwritten digit classification
task~\cite{globerson-ICML06,LeCun95}, as with the evasion attacks in
Section~\ref{sect:exp-evasion-test-time}.  In particular, we focus
here on the following two-class sub-problems: 7 vs. 1; 9 vs.  8; 4 vs.
0.
%Each digit in the MNIST dataset is properly normalized and represented as a grayscale image of $28 \times 28$ pixels. In particular, each pixel is ordered in a raster-scan and its value is directly considered as a feature. The overall number of features is $d = 28 \times 28 = 784$. We normalized each feature (pixel value) $x \in [0,1]^{d}$ by dividing its value by $255$.
Each digit is normalized as described in
Section~\ref{sect:toy-example}.  We consider again a linear SVM with
$C=1$. We randomly sample a training and a validation data of $100$
and $500$ samples, respectively, and retain the complete testing data
given by MNIST for $\mathcal D_{\rm ts}$. Although it varies for each
digit, the size of the testing data is about 2000 samples per class
(digit).

\begin{figure*}[tb]
\begin{center}
\includegraphics[width=0.75\textwidth]{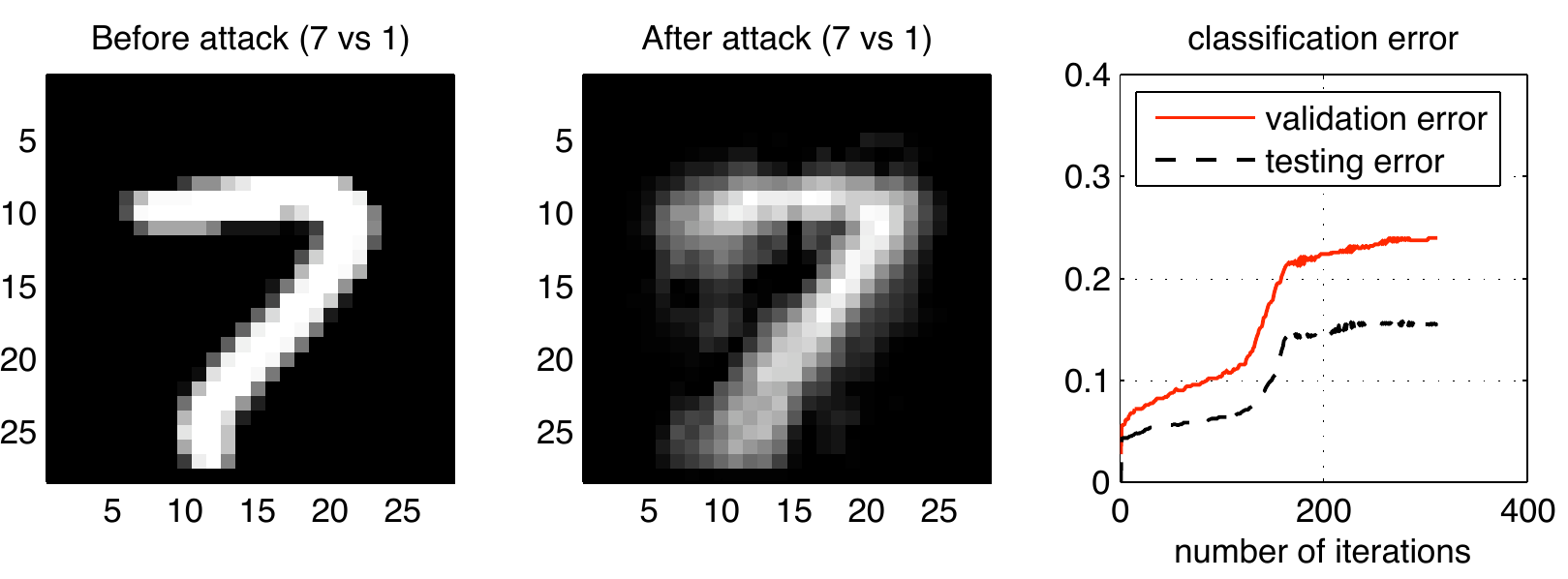}\\
\includegraphics[width=0.75\textwidth]{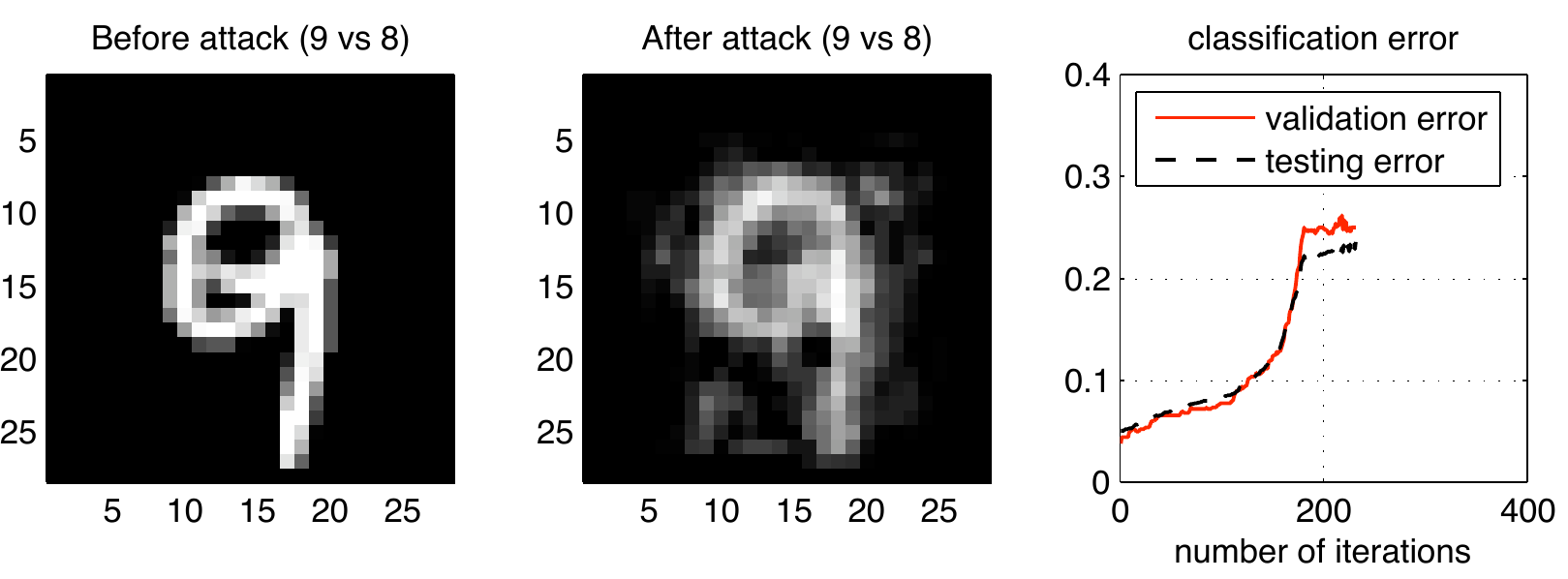}\\
\includegraphics[width=0.75\textwidth]{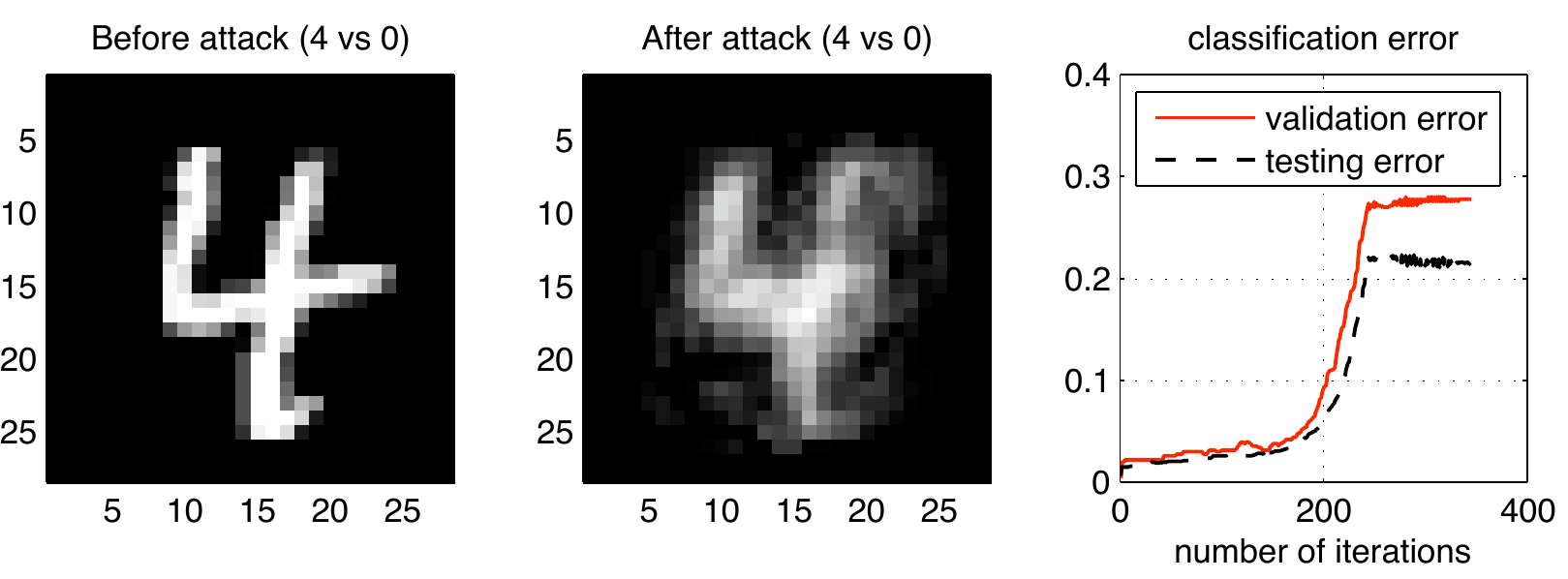}
\caption{Modifications to the initial (mislabeled) attack point
  performed by the proposed attack strategy, for the three considered
  two-class problems from the MNIST dataset. The increase in
  validation and testing errors across different iterations is also
  reported.}
\label{fig:exp-single-digits}
\end{center}
\end{figure*}

Figure~\ref{fig:exp-single-digits} shows the effect of \emph{single}
attack points being optimized by our descent method.  The leftmost
plots of each row show the example of the attacked class used as
starting points in our algorithm. The middle plots show the final
attack point. The rightmost plots depict the increase in the
validation and testing errors as the attack progresses. For this
experiment we run the attack algorithm $5$ times by re-initializing the
gradient ascent procedure, and we retain the best result.

Visualizing the attack points reveals that these attacks succeed by
blurring the initial prototype to appear more like examples of the
attacking class. In comparing the initial and final attack points, we
see that the bottom segment of the $7$ straightens to resemble a $1$,
the lower segment of the $9$ is rounded to mimicking an $8$, and
\emph{ovular} noise is added to the outer boundary of the $4$ to make
it similar to a $0$. These blurred images are thus consistent with
one's natural notion of visually confusing digits.

The rightmost plots further demonstrate a striking increase in error
over the course of the attack. In general, the validation error
overestimates the classification error due to a smaller sample size.
Nonetheless, in the exemplary runs reported in this experiment, a
\emph{single} attack data point caused the classification error to
rise from initial error rates of 2--5\% to 15--20\%.
%BAT: comment on point 3
Since our initial attack points are obtained by flipping the label of
a point in the attacked class, the errors in the first iteration of
the rightmost plots of Figure~\ref{fig:exp-single-digits} are caused
by single random label flips.
%
%This finding underscores the vulnerability of the SVM to poisoning attacks. 
This confirms that our attack can achieve significantly higher error
rates than random label flips, and underscores the vulnerability of
the SVM to poisoning attacks.

\begin{figure}[tbhp]
\begin{center}
\includegraphics[width=0.49\textwidth]{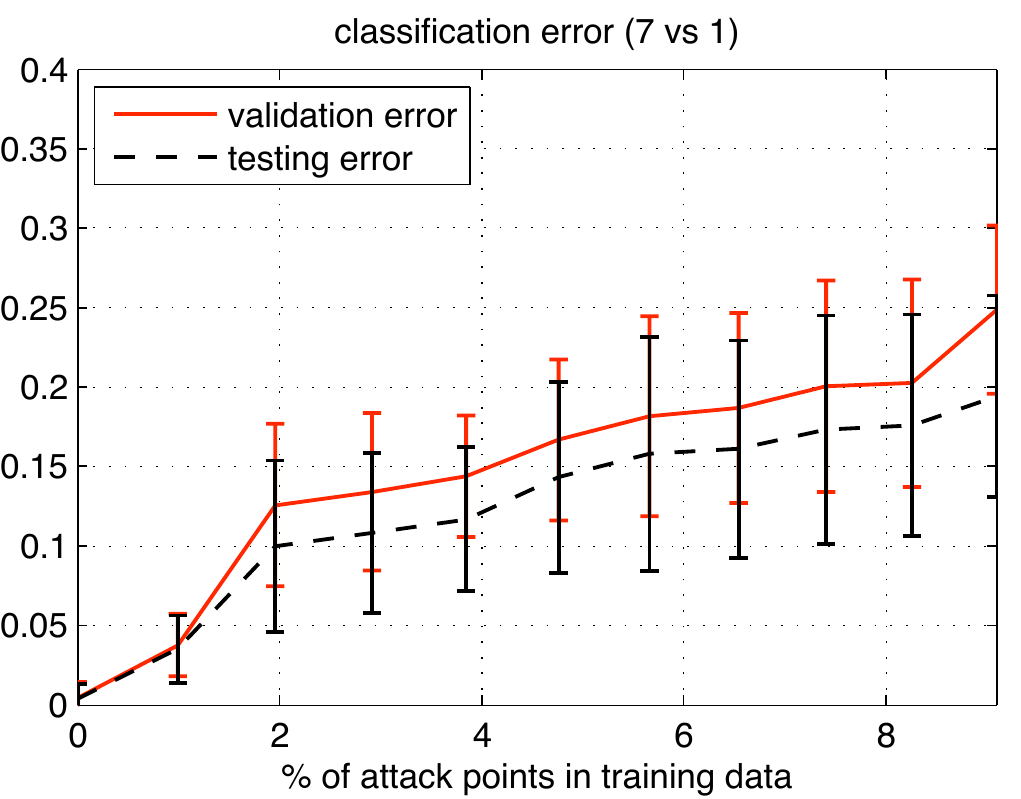}
\includegraphics[width=0.49\textwidth]{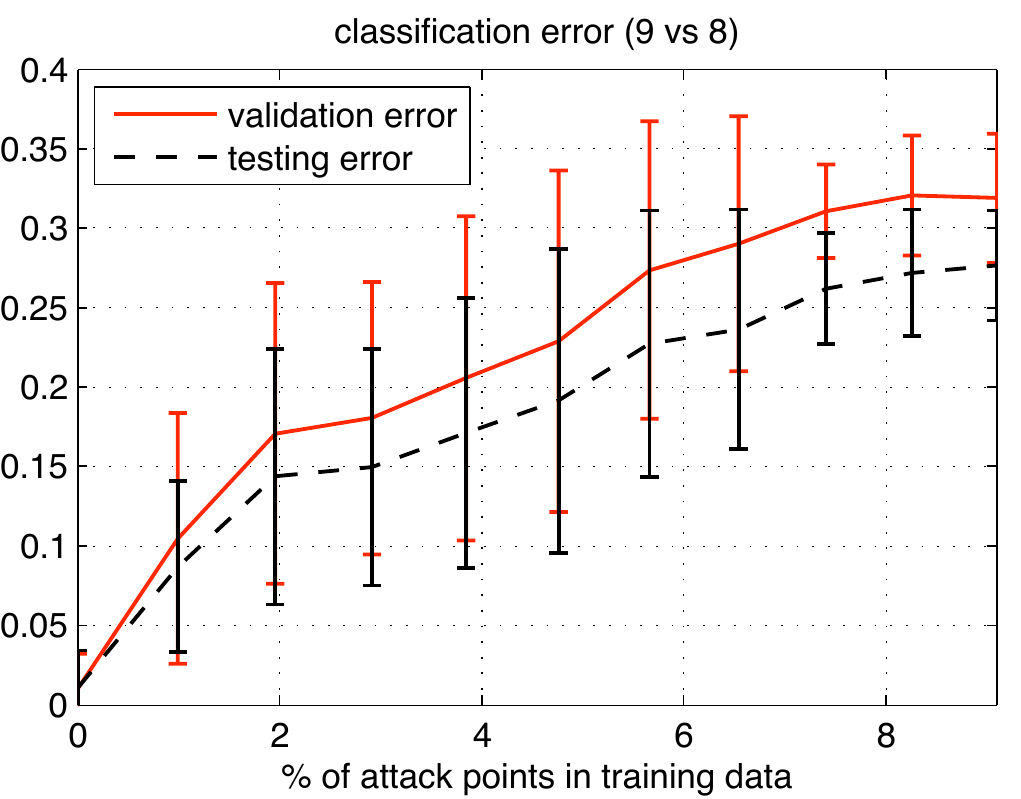}
\includegraphics[width=0.49\textwidth]{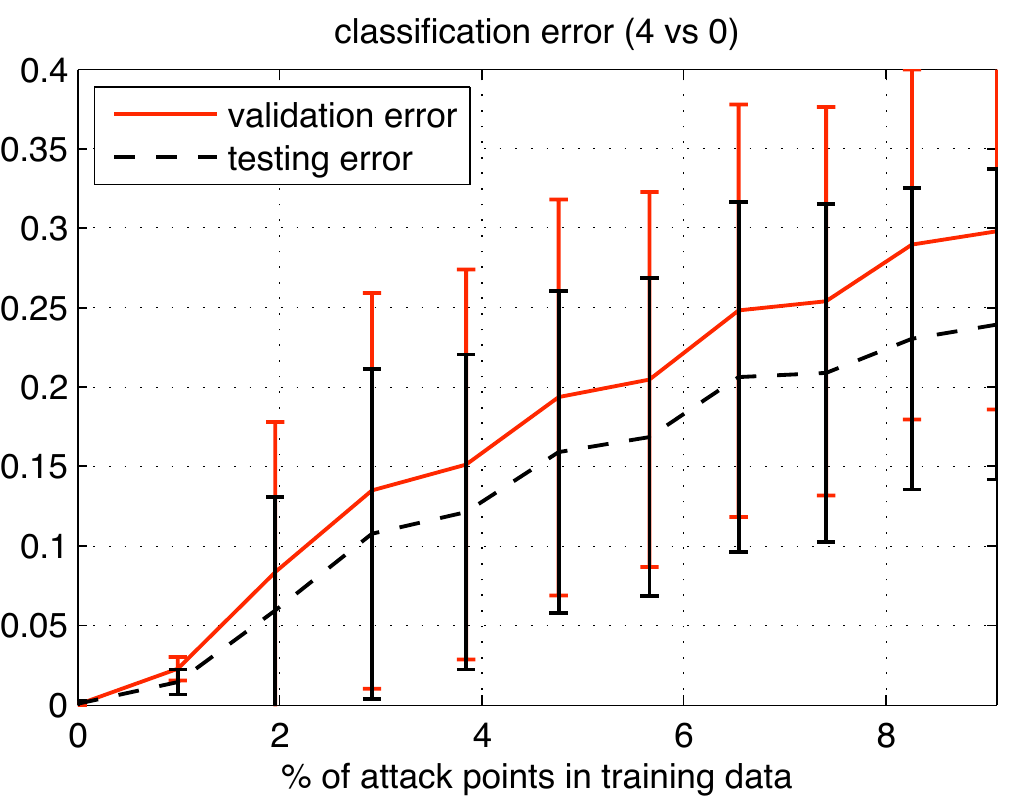}
\end{center}
\caption{Results of the multi-point, multi-run experiments on the
  MNIST dataset. In each plot, we show the classification errors due
  to poisoning as a function of the percentage of training
  contamination for both the validation (red solid line) and testing
  sets (black dashed line). The top-left plot is for the $7$ vs.$1$
  task, the top-right plot is for the $9$ vs. $8$ task, and the
  bottom-middle plot is for the $4$ vs. $0$ task.}
\label{fig:exp-multi-digits}
\end{figure}

The latter point is further illustrated in a multiple point, multiple
run experiment presented in Figure~\ref{fig:exp-multi-digits}. For
this experiment, the attack was extended by repeatedly injecting
attack points into the same class and averaging results over multiple
runs on randomly chosen training and validation sets of the same size
(100 and 500 samples, respectively). These results exhibit a steady
rise in classification error as the percentage of attack points in the
training set increases. The variance of the error is quite high, which
can be attributed to the relatively small sizes of the training and
validation sets. Also note that, in this experiment, to reach an error
rate of 15--20\%, the adversary needs to control at least 4--6\% of
the training data, unlike in the single point attacks of
Figure~\ref{fig:exp-single-digits}. This is because
Figure~\ref{fig:exp-single-digits} displays the best single point
attack from five restarts whereas here initial points are selected
without restarts.
%
%The reason is that we consider
%here a single initialization of the attack algorithm, instead of
%running the algorithm $5$ times and retaining the best result (as is done
%in the former experiment).

\subsection{Discussion}
\label{sect:poisoning-discussions}

The poisoning attack presented in this section, summarized from our
previous work in~\cite{biggio12-icml}, is a first step toward the
security analysis of SVM against training data attacks. Although our
gradient ascent method is not optimal, it attains a surprisingly large
impact on the SVM's classification accuracy.

Several potential improvements to the presented method remain to
be explored in future work.
For instance, one may investigate the effectiveness of such an attack with surrogate data, that is, when the training data is not known to the adversary, who may however collect samples drawn from the same distribution to learn a classifier's copy (similarly to the limited knowledge case considered in the evasion attacks of Section~\ref{sect:evasion}).
Another improvement may be to consider the simultaneous optimization of multi-point
attacks, although we have already demonstrated that greedy, sequential single-point
attacks may be rather successful. 

An interesting analysis of the SVM's vulnerability to poisoning
suggested from this work is to consider the attack's impact under loss
functions other than the hinge loss.  It would be especially
interesting to analyze \emph{bounded} loss functions, like the ramp
loss, since such losses are designed to limit the impact of any
single (attack) point on the outcome.  On the other hand,
while these losses may lead to improved security to poisoning, they
also make the SVM's optimization problem non-convex, and, thus, more
computationally demanding. This may be viewed as another trade-off
between computational complexity of the learning algorithm and
security.

An important practical limitation of the proposed method is the
assumption that the attacker controls the labels of injected
points. Such assumptions may not hold if the labels
are assigned by trusted sources such as humans, \eg,
anti-spam filters use their users'
labeling of messages as ground truth. Thus, although an attacker can
send arbitrary messages, he cannot guarantee that they will have the
labels necessary for his attack. This imposes an additional requirement
that the attack data must satisfy certain side constraints to
fool the labeling oracle. Further work is needed to understand and incorporate these
potential side constraints into attacks.

\section{Privacy Attacks against SVMs}
\label{sect:privacy}

We now consider a third scenario in which the attacker's goal is to
affect a breach of the training data's confidentiality. We review our recent
work~\cite{PrivateSVM} deriving mechanisms for releasing SVM
classifiers trained on privacy-sensitive data while maintaining the
data's privacy. Unlike previous sections, our focus here lies primary
in the study of countermeasures, while we only briefly consider
attacks in the context of lower bounds.  We adopt the formal framework
of Dwork~\etal~\cite{DiffPrivacy}, in which a randomized mechanism is
said to preserve \emph{\dppriv-differential privacy}, if the
likelihood of the mechanism's output changes by at most \dppriv when a
training datum is changed arbitrarily (or even removed). The power of
this framework, which gained near-universal favor after its
introduction, is that it quantifies privacy in a rigorous way and
provides strong guarantees even against powerful adversaries
with knowledge of almost all of the training data, knowledge of the
mechanism (barring its source of randomness), arbitrary access to the
classifier output by the mechanism, and the ability to manipulate
almost all training data prior to learning.

This section is organized as follows. In Section~\ref{sect:privacy-model}
we outline our model of the adversary, which makes only weak assumptions.
Section~\ref{sect:privacy-algo} provides background on differential privacy,
presents a mechanism for training and releasing privacy-preserving
SVMs---essentially a countermeasure to many privacy attacks---and provides
guarantees on differential privacy and also utility (\eg, controlling the
classifier's accuracy). We then briefly touch on existing approaches for
evaluation via lower bounds and
discuss other work and open problems in Section~\ref{sect:privacy-disc}.

\subsection{Modeling the Adversary}
\label{sect:privacy-model}

We first apply our framework to define the threat model for defending
against privacy attacks within the broader context of differential
privacy. We then focus on specific countermeasures in the form of
modifications to SVM learning that provide differential privacy.

\subsubsection{Adversary's Goal}

The ultimate goal of the attacker in this section is to determine features
and/or the label of an individual training datum. The overall approach of the
adversary towards this goal, is to inspect (arbitrary numbers of) test-time
classifications made by a released classifier trained on the data, or by
inspecting the classifier directly. The definition of differential
privacy, and the particular mechanisms derived here, can be modified for 
related goals of determining properties of several training data; we
focus on the above conventional case without loss of generality.

\subsubsection{Adversary's Knowledge}

As alluded to above, we endow our adversary with significant knowledge of 
the learning system, so as to derive countermeasures that can withstand
very strong attacks. Indeed the notion of differential privacy, as opposed to
more syntactic notions of privacy such as \emph{$k$-anonymity}~\cite{kanony},
was
inspired by decades-old work in cryptography that introduced mathematical
formalism to an age-old problem, yielding significant practical success.
%DM: elluded -> alluded
Specifically, we consider a scenario in which the adversary has complete
knowledge of the raw input feature representation, the learning algorithm (the
entire mechanism including the form of randomization it introduces, although
not the source of randomness) and the form of its decision function (in this
case, a thresholded SVM), the learned classifier's parameters (the kernel/feature mapping, primal weight vector, and
bias term), and arbitrary/unlimited feedback from the deployed classifier
\knowref[know-2]{know-5}. We grant the attacker near
complete knowledge of the training set \knowref{know-1}: the attacker may
have complete knowledge of all but one training datum, for which she has no
knowledge of input feature values or its training label, and it is these attributes
she wishes to reveal. For simplicity of exposition, but without loss of 
generality, we assume this to be the last datum in the training sample.

\subsubsection{Adversary's Capability}

Like our assumptions on the attacker's knowledge, we impose weak 
limitations on the adversary's capability.
We assume an adversary that can manipulate both training and test data
\capref{cap-1}, although the latter is subsumed by 
the attacker's complete knowledge of the decision function and
learned parameters---\eg, she may implement her own classifier and execute
it arbitrarily, or she may submit or manipulate test points presented to
a deployed classifier.

Our attack model makes no assumptions about the origins of the
training or test data. The data need not be sampled
independently or even according to a distribution---the definition of
differential privacy provided below makes worst-case assumptions about
the training data, and again the test data could be arbitrary. Thus
the adversary may have arbitrary capability to modify class priors,
training data features and labels~\capref[cap-2]{cap-4} except that
the adversary attacking the system may not directly modify the targeted
training datum because she does not have
knowledge of it. That said, however, differential privacy makes
worst-case (no distributional) assumptions about the datum and thus
one could consider even this data point as being adversarially
manipulated by nature (\ie, nature does not collude with the
attacker to share information about the target training datum,
but that may collude to facilitate a privacy breach by selecting a
``convenient'' target datum).

\subsubsection{Attack Strategy}

While no practical privacy attacks on SVMs have been explored in the
past---an open problem discussed in
Section~\ref{sect:privacy-disc}---a general approach would be to
approximate the inversion of the learning map on the released SVM
parametrization (either primal weight vector, or dual variables)
around the known portion of the training data. In practice this could
be achieved by taking a similar approach as done in
Section~\ref{sect:poisoning} whereby an initial guess of a missing
training point is iterated on by taking gradient steps of the
differential in the SVM parameter vector with respect to the missing
training datum. An interpretation of this approach is one of
simulation: to guess a missing training datum, given access to the
remainder of the training set and the SVM solution on all the data,
simulate the SVM on guesses for the missing datum, updating the
guesses in directions that appropriately shift the intermediate
solutions. As we discuss briefly in the sequel, theoretical lower
bounds on achievable privacy relate to attacks in pathological cases.

\subsection{Countermeasures with Provable Guarantees}
\label{sect:privacy-algo}

Given an adversary with such strong knowledge and capabilities as
described above, it may seem difficult to provide effective
countermeasures particularly considering the complication of abundant
access to side information that is often used in publicized privacy
attacks~\cite{break-netflix,kanony}. However, the crux that makes
privacy-preservation under these conditions possible lies in the fact
that the learned quantity being released is an aggregate
statistic of the sensitive data; intuitively the more data being
aggregated, the less sensitive a statistic should be to changes or
removal of any single datum. We now present results from our recent work
that quantifies this effect~\cite{PrivateSVM}, within the framework of
\emph{differential privacy}.

\subsubsection{Background on Differential Privacy}

We begin by recalling the key definition due to Dwork \etal~\cite{DiffPrivacy}.
First, for any training set $\mathcal D=\{(\mathbf x_{i}, y_{i}) \}_{i=1}^{n}$
denote
set $\mathcal D'$ to be a \emph{neighbor} of $\mathcal D$ (or $\mathcal D'\sim \mathcal D$) if $\mathcal D'=\{(\mathbf x_{i}, y_{i}) \}_{i=1}^{n-1}\cup \{(\mathbf x'_n, y'_n)\}$ where $(\mathbf x_n, y_n)\neq(\mathbf x'_n, y'_n)$.
In the present context, differential privacy is a desirable property of
\emph{learning maps}, which maps a training set 
$\{(\mathbf x_{i}, y_{i}) \}_{i=1}^{n}$ to a continuous discriminant
function of the form $g: \mathcal{X} \to \R$---here a learned SVM---in some
space of functions, \hypotheses. We say that a \emph{randomized}\footnote{That
is, the learning map's output is not a deterministic function of the training
data. The probability in the definition of differential privacy is due to this
randomness. Our treatment here is only as complex as necessary, but to be
completely general, the events in the definition should be on measurable sets $G\subset \hypotheses$ rather than individual $g\in\hypotheses$.} learning map \mech\ preserves \dppriv-differential privacy if for
all datasets $\mathcal D$, all neighboring sets $\mathcal D'$ of $\mathcal D$, and all possible
functions
$g\in \hypotheses$, the following relation holds
\begin{eqnarray*}
\Pr(\mech(\mathcal D) = g) &\leq& \exp(\dppriv) \Pr(\mech(\mathcal D') = g)\enspace.
\end{eqnarray*}
Intuitively, if we initially fix a training set and neighboring training set,
differential privacy simply says that the two resulting distributions induced
on the learned functions are point-wise close---and closer for smaller \dppriv.
For a patient deciding whether to submit her datum to a training set for a 
cancer detector, differential privacy means that the learned classifier will reveal little information about that datum.
% BLAINE:  so why do it then?
%knowing that the detector preserves differential privacy means
%that the particular value of her individual datum 
%does not significantly affect the training process. 
Even an adversary with access to the
inner-workings of the learner, with access to all other patients' data,
and with the ability to guess-and-simulate the learning process repeatedly
with various possible values of her datum, cannot reverse engineer her datum
from the classifier released by the hospital because the adversary cannot
distinguish the classifier distribution on one training set, from that on
neighboring sets. Moreover, variations of this definition (which do not
significantly affect the presented results) allow for neighboring databases to
be defined as those missing a datum; or having several varying data, not just
a single one.

For simplicity of exposition, we drop the explicit bias term $b$ from
our SVM learning process and instead assume that the data feature
vectors are augmented with a unit constant, and that the resulting
additional normal weight component corresponds to the bias. This is an
equivalent SVM formulation that allows us to focus only on the normal's
weight vector.

A classic route to establish differential privacy is to define a
randomized map \mech that returns the value of a deterministic,
non-random \npmech plus a noise term. Typically, we use an exponential
family in a term that matches an available Lipschitz condition
satisfied by \npmech: in our case, for learning maps that return
weight vectors in $\R^d$, we aim to measure \emph{global sensitivity}
of \npmech via the $L_1$ norm as
\begin{eqnarray*}
\Delta(\npmech) &=& \max_{\mathcal D,\mathcal D'\sim \mathcal D} \left\|\npmech(\mathcal D) - \npmech(\mathcal D')\right\|_1\enspace.
\end{eqnarray*}
With respect to this sensitivity, we can easily prove that the randomized mechanism
\begin{eqnarray*}
\mech(\mathcal D) &=& \npmech(\mathcal D) + Laplace(0, \Delta(\npmech) / \beta)\enspace,
\end{eqnarray*}
is $\beta$-differential private.\footnote{Recall that the zero-mean
multi-variate Laplace distribution with scale parameter $s$ has density proportional to
$\exp(-\|\mathbf x\|_1/s)$.}
The well-established proof technique~\cite{DiffPrivacy} follows from the definition of
the Laplace distribution involving the same norm as used in our measure of global
sensitivity, and the triangle inequality: for any training set $\mathcal D$, $\mathcal D'\sim \mathcal D$,
response $g\in\hypotheses$, and privacy parameter $\beta$
\begin{eqnarray*}
\frac{\Pr(\mech(\mathcal D) = g)}{\Pr(\mech(\mathcal D') = g)}
&=& \frac{\exp\left(\left\|\npmech(\mathcal D') - g\right\|_1 \beta / \Delta(\npmech)\right)}{\exp\left(\left\|\npmech(\mathcal D) - g\right\|_1 \beta / \Delta(\npmech)\right)} \\
&\leq& \exp\left(\left\|\npmech(\mathcal D') - \npmech(\mathcal D)\right\|_1 \beta / \Delta(\npmech)\right) \\
&\leq& \exp(\beta) \enspace.
\end{eqnarray*}
We take the above route to develop a differentially-private SVM. As
such, the onus is on calculating the SVM's global sensitivity, $\Delta(\npmech)$.

\subsubsection{Global Sensitivity of Linear SVM}

Unlike much prior work applying the ``Laplace mechanism'' to achieving
differential privacy, in which studied estimators are often decomposed as
linear functions of data~\cite{SuLQ},
measuring the sensitivity of the SVM appears to be non-trivial
owing to the non-linear influence an individual training datum may have on
the learned SVM. However, perturbations of the training data were
studied by the learning-theory community in the context of 
\emph{algorithmic stability}: there the goal is to establish bounds on
classifier risk, from stability of the learning map, as opposed to
leveraging combinatorial properties of the hypothesis class (\eg, the
VC dimension, which is not always possible to control, and for the RBF kernel SVM is
infinite)~\cite{SVMbook}.
In recent work~\cite{PrivateSVM}, we showed how these existing stability
measurements for the SVM can be adapted to provide the following $L_1$-global
sensitivity bound.

\begin{lemma}
Consider SVM learning with a kernel corresponding to linear SVM in a feature
space with finite-dimension $F$ and $L_2$-norm bounded\footnote{That is 
$\forall \mathbf x$, $k(\mathbf x, \mathbf x)\leq\kappa^2$; \eg\ for the
RBF the norm is uniformly unity $\kappa=1$; more generally, 
we can make the standard assumption that the data lies within some $\kappa$
$L_2$-ball.} by $\kappa$, with hinge loss (as used throughout this chapter), and chosen parameter
$C>0$. Then the $L_1$ global sensitivity of the resulting normal weight vector
is upper-bounded by $4C\kappa\sqrt{F}$.
\end{lemma}

We omit the proof, which is available in the original paper~\cite{PrivateSVM} and which follows
closely the previous measurements for algorithmic stability. We note that
the result extends to any convex Lipschitz loss.

\begin{algorithm}[tb]
  \caption{Privacy-preserving SVM}
  \label{alg:private}
  \textbf{Input:} $\mathcal D$ the training data; $C>0$ soft-margin parameter; kernel $k$ inducing a feature space with finite dimension $F$ and $\kappa$-bounded $L_2$-norm; privacy parameter $\beta>0$.\\
  \textbf{Output:} learned weight vector $\mathbf w$ .
  
  \begin{algorithmic}[1]
    \STATE{ $\mathbf{\hat w} \leftarrow $ learn an SVM with parameter $C$ and kernel $k$ on data $\mathcal D$.}
    \STATE{ $\mathbf{\mu} \leftarrow$ draw i.i.d. sample of $F$ scalars from $Laplace\left(0, \frac{4 C\kappa\sqrt{F}}{\dppriv}\right)$.}
    \STATE{\textbf{return:} $\mathbf{w} = \mathbf{\hat{w}} + \mathbf \mu$}
  \end{algorithmic}
\end{algorithm}

\subsubsection{Differentially-Private SVMs}

So far we have established that Algorithm~\ref{alg:private}, which
learns an SVM and returns the resulting weight vector with added
Laplace noise, preserves $\dppriv$-differential privacy. More noise is
added to the weight vector when either (i) a higher degree of privacy
is desired (smaller $\beta$), (ii) the SVM fits closer to the data
(higher $C$) or (iii) the data is more distinguishable (higher
$\kappa$ or $F$---the curse of dimensionality). Hidden in the above is
the dependence on $n$: typically we take $C$ to scale like $1/n$
to achieve consistency in which case we see that noise decreases
with larger training data---akin to less individual influence---as
expected~\cite{PrivateSVM}.

Problematic in the above approach, is the destruction to utility due to  
preserving differential privacy. One approach to quantifying this effect,
involves bounding the following notion of utility~\cite{PrivateSVM}. We say a
privacy-preserving learning map \mech has $(\dpacc,\dpconf)$-utility with 
respect to non-private map \npmech if for all training sets $\mathcal D$,
\begin{eqnarray*}
\Pr\left(\left\|\mech(\mathcal D) - \npmech(\mathcal D) \right\|_\infty \leq \dpacc\right) & \geq & 1 - \dpconf\enspace.
\end{eqnarray*}
The norm here is in the function space of continuous discriminators, learned
by the learning maps, and is the point-wise $L_\infty$ norm which corresponds
to $\|g\|_\infty = \sup_{\mathbf x} |g(\mathbf x)|$---although for technical
reasons we will restrict the supremum to be over a set $\mathcal M$ to be specified later.
Intuitively, this indicates that the continuous predictions of the learned private classifier are close to
those predictions of the learned non-private classifier, for all test points
in $\mathcal M$,
with high probability (again, in the randomness due to the private mechanism).
This definition draws parallels with PAC learnability, and in certain
scenarios is strictly stronger than requiring that the private learner
achieves good risk (\ie, PAC learns)~\cite{PrivateSVM}. Using the Chernoff tail
inequality and known moment-generating functions, we establish the
following bound on the utility of this private SVM~\cite{PrivateSVM}.

\begin{theorem}
The $\dppriv$-differentially-private SVM of Algorithm~\ref{alg:private} achieves
$(\dpacc, \dpconf)$-utility with respect to the non-private SVM run with
the same $C$ parameter and kernel, for $0<\delta<1$ and
\begin{eqnarray*}
\epsilon &\geq& 8 C\kappa\Phi\sqrt{F}\left(F + \log\frac{1}{\delta}\right) / \beta\enspace,
\end{eqnarray*}
where the set $\mathcal M$ supporting the supremum in the definition of utility
is taken to be the pre-image of the feature mapping on the $L_\infty$ ball of radius
$\Phi > 0$.\footnote{Above we previously bounded the $L_2$ norms of points
in features space by $\kappa$, the additional bound on the $L_\infty$ norm here is for convenience
and is standard practice in learning-theoretic results.}
\end{theorem}

As expected, the more confidence $\dpconf$ or privacy $\dppriv$
required, the less accuracy is attainable. Similarly, when the
training data is fitted more tightly via higher $C$, or when the data
is less tightly packed for higher $\kappa, \Phi, F$, less accuracy is
possible. Note that like the privacy result, this result can hold for
quite general loss functions.

\subsection{Discussion}
\label{sect:privacy-disc}

In this section, we have provided a summary of our recent results on 
strong counter-measures to privacy attacks on the SVM.
We have shown how, through controlled addition of noise, SVM learning in finite-dimensional feature spaces
can provide both privacy and utility guarantees. 
These results can be extended to certain translation-invariant
kernels including the infinite-dimensional RBF~\cite{PrivateSVM}.
This extension borrows a technique from large-scale
learning where finding a dual solution of the SVM for large training data size $n$ is
infeasible. Instead, a primal SVM problem is solved using a random kernel
that uniformly approximates the desired kernel. Since the approximating kernel
induces a feature mapping of relatively small, finite dimensions, the primal
solution becomes feasible. For 
privacy preservation, we use the same primal approach but with this new
kernel. Fortunately, the distribution of the approximating kernel is 
independent of the training data, and thus we can reveal
the approximating kernel without sacrificing privacy.
Likewise the uniform approximation of the kernel composes with the utility
result here to yield an analogous utility guarantee for translation-invariant
kernels.

While we demonstrated here a mechanism for private SVM learning with
upper bounds on privacy and utility, we have previously also studied
lower bounds that expose limits on the achievable utility of
\emph{any} learner that provides a given level of differential
privacy.  Further work is needed to sharpen these results. In a sense,
these lower bounds are witnessed by pathological training sets and
perturbation points and, as such, serve as attacks in pathological
(unrealistic) cases. Developing practical attacks on the privacy of an
SVM's training data remains unexplored.

Finally, it is important to note that alternate approaches to
differentially-private SVMs have been explored by others. Most notable
is the work (parallel to our own) of Chaudhuri
\etal~\cite{PrivateERM}. Their approach to finite-dimensional feature
mappings is, instead of adding noise to the primal solution, to add
noise to the primal objective in the form of a dot product of the
weight with a random vector. Initial experiments show their approach
to be very promising empirically, although it does not allow for
non-differentiable losses like the hinge loss.

\section{Concluding Remarks}
\label{sect:conclusions}
In security applications like malware detection, intrusion detection, and spam filtering, SVMs may be attacked through patterns that can either evade detection (evasion), mislead the learning algorithm (poisoning), or gain information about their internal parameters or training data (privacy violation). In this chapter, we demonstrated that these attacks are feasible and constitute a relevant threat to the security of SVMs, and to machine learning systems, in general. 

\textbf{Evasion}. We proposed an \emph{evasion} algorithm against SVMs
with differentiable kernels, and, more generally, against classifiers
with differentiable discriminant functions. We investigated the
attack's effectiveness in perfect and limited knowledge settings. In
both cases, our attack simulation showed that SVMs (both linear and
RBF) can be evaded with high probability after a few modifications to
the attack patterns. Our analysis also provides some general hints for
tuning the classifier's parameters (\eg, the value of $\gamma$ in SVMs
with the RBF kernel) and for improving classifier security. For
instance, if a classifier tightly \emph{encloses} the legitimate
samples, the adversary's samples must closely mimic legitimate samples
to evade it, in which case, if such exact mimicry is still possible, it suggests an
inherent flaw in the feature representation.

\textbf{Poisoning}. We presented an algorithm that allows the
adversary to find an attack pattern whose addition to the training set
maximally decreases the SVM's classification accuracy. We found that
the increase in error over the course of attack is especially
striking. A single attack data point may cause the classification
error to rise from the initial error rates of 2--5\% to 15--20\%. This
confirms that our attack can achieve significantly higher error rates
than random label flips, and underscores the vulnerability of the SVM
to poisoning attacks. As a future investigation, it may be of interest
to analyze the effectiveness of poisoning attacks against non-convex
SVMs with bounded loss functions, both empirically and theoretically,
since such losses are designed to limit the impact of any single
(attack) point on the resulting learned function.  This has been also
studied from a more theoretical perspective in \cite{christmann04},
exploiting the framework of Robust Statistics
\cite{hampel86,maronna06}.  A similar effect is obtained by using
bounded kernels (\eg, the RBF kernel) or bounded feature values.

\textbf{Privacy}. We developed an SVM learning algorithm that preserves
\emph{differential privacy}, a formal framework for quantifying the threat of a potential training
set privacy violation incurred by releasing learned classifiers. Our 
mechanism involves adding Laplace-distributed noise to the SVM weight
vector with a scale that depends on the algorithmic stability of the SVM and
the desired level of privacy. In addition to presenting a formal guarantee that our
mechanism preserves privacy, we also provided bounds on the utility of the 
new mechanism, which state that the privacy-preserving classifier makes
predictions that are point-wise close to those of the non-private SVM, with
high probability. Finally we discussed potential approaches for attacking
SVMs' training data privacy, and known approaches to differentially-private
SVMs with (possibly infinite-dimensional feature space) translation-invariant
kernels, and lower bounds on the fundamental limits on utility for private
approximations of the SVM.

\begin{acknowledgement}
This work has been partly supported by the project CRP-18293 funded by Regione Autonoma della Sardegna, L.R. 7/2007, Bando 2009, and by the project ``Advanced and secure sharing of multimedia data over social networks in the future Internet'' (CUP F71J1100069
0002) funded by the same institution.
Davide Maiorca gratefully acknowledges Regione Autonoma della Sardegna for the financial support of his PhD scholarship (P.O.R. Sardegna F.S.E. Operational Programme of the Autonomous Region of Sardinia, European Social Fund 2007-2013 - Axis IV Human Resources, Objective l.3, Line of Activity l.3.1.). Blaine Nelson thanks the Alexander von
Humboldt Foundation for providing additional financial support. The opinions expressed in this chapter are solely
those of the authors and do not necessarily reflect the opinions of
any sponsor.
\end{acknowledgement}

%the Springer BibTeX styles "spbasic.bst", "spmpsci.bst", "spphys.bst"
%\bibliographystyle{spmpsci}
%\bibliography{bibDB,biblio,bib-ic}

\end{document}